\documentclass[11pt,a4paper,logo,nonumbering]{style}
\usepackage[numbers,sort&compress]{natbib}
\usepackage{fontawesome5}

\usepackage{amsmath,amsfonts,bm}

\def\eqref#1{equation~\ref{#1}}

\def\1{\bm{1}}

\DeclareMathAlphabet{\mathsfit}{\encodingdefault}{\sfdefault}{m}{sl}
\SetMathAlphabet{\mathsfit}{bold}{\encodingdefault}{\sfdefault}{bx}{n}

\newcommand{\nConfigsMeasured}{48}
\newcommand{\nRunsMeasured}{240}

\newcommand{\nMissingCons}{7}
\newcommand{\nMissingPlaus}{4}

\newcommand{\ClaudeMeanS}{47.86}
\newcommand{\CodexMeanS}{8.47}
\newcommand{\ClaudeMeanC}{0.814}
\newcommand{\CodexMeanC}{0.136}
\newcommand{\ClaudeMeanQ}{0.513}
\newcommand{\CodexMeanQ}{0.616}
\newcommand{\BriefOneC}{0.647}
\newcommand{\BriefFiveC}{0.284}

\newcommand{\DesignPairs}{990}
\newcommand{\DesignUntiedPairs}{978}
\newcommand{\DesignDiscordantPairs}{545}
\newcommand{\DesignDiscordantPercent}{55.7}
\newcommand{\DesignCQrho}{-0.033}
\newcommand{\DesignCSrho}{0.956}
\newcommand{\DesignCQFullrho}{-0.025}
\newcommand{\DesignWeightMinrho}{0.825}
\newcommand{\DesignWeightMaxrho}{0.990}
\newcommand{\DesignMeanGap}{0.063}

\newcommand{\nBfiveLower}{47}

\newcommand{\imageModel}{\texttt{gemini-3}}
\newcommand{\videoModel}{\texttt{seedance-2.0-mini}}

\usepackage{hyperref}
\hypersetup{pdftitle={VAHBench: A Harness Benchmark for Agentic Long-Video Generation},pdfauthor={Hongbo Zhang, Liuyang Song, Hang Zhou, Xinchen Liu, Wei He, Kai Han, Yunhe Wang}}
\usepackage{url}
\usepackage{graphicx}
\usepackage{booktabs}
\usepackage{amsmath}
\usepackage{amssymb}
\usepackage{xcolor}
\usepackage{multirow}
\usepackage{tikz}
\usetikzlibrary{arrows.meta,positioning,calc}
\usepackage[most]{tcolorbox}
\usepackage{enumitem}
\usepackage{array}
\usepackage{longtable}
\usepackage{tabularx}
\usepackage{placeins}
\usepackage{float}
\usepackage{makecell}
\usepackage{textcomp}
\usepackage{pifont}
\usepackage{colortbl}
\definecolor{TableBlue}{HTML}{E5F1FA}
\definecolor{TableYellow}{HTML}{FFF3C9}
\definecolor{TableRule}{HTML}{668EAD}
\AtBeginEnvironment{table}{\arrayrulecolor{TableRule}}
\DeclareMathSizes{7.5}{7.5}{5.5}{5}
\newcommand{\cmark}{\ding{51}}
\newcommand{\xmark}{\ding{55}}


\renewcommand{\topfraction}{0.80}
\renewcommand{\textfraction}{0.15}
\graphicspath{{figures/}}

\newtcolorbox{promptbox}[2]{%
  breakable, enhanced, arc=2.6mm, boxrule=0.5pt,
  colback=TableBlue!35!white, colframe=TableRule,
  left=3mm, right=3mm, top=1.6mm, bottom=2mm,
  fonttitle=\small, coltitle=black, colbacktitle=TableBlue,
  title={\textbf{#1}\hfill\footnotesize #2}}
\newcommand{\promptsec}[1]{%
  \par\vspace{1.2mm}\noindent{\footnotesize\scshape #1}\par\nobreak\vspace{0.3mm}\nobreak}
\newcommand{\promptpar}[1]{\par\noindent{\footnotesize #1}\par\vspace{0.4mm}}
\newlist{promptlist}{itemize}{1}
\setlist[promptlist]{label=\raisebox{0.2ex}{\tiny$\bullet$}, leftmargin=4.2mm,
  itemsep=0.2mm, topsep=0.5mm, parsep=0pt, partopsep=0pt, beginpenalty=10000, before=\footnotesize}

\newcommand{\bench}{VAHBench}

\title{\centering \bench: A Harness Benchmark for Agentic Long-Video Generation}

\author{Hongbo Zhang, Liuyang Song, Hang Zhou, Xinchen Liu, Wei He, Kai Han, Yunhe Wang}
\makeatletter
\renewcommand{\@author}{%
  \begin{minipage}{\linewidth}
  \centering\normalfont\fontsize{11}{15}\selectfont
  \href{https://openreview.net/profile?id=~Hongbo_Zhang11}{\textcolor{black}{Hongbo Zhang}}\textsuperscript{1,2,\textdagger},
  \href{https://openreview.net/profile?id=~Liuyang_Song3}{\textcolor{black}{Liuyang Song}}\textsuperscript{1,2,\textdagger},
  \href{https://openreview.net/profile?id=~Hang_Zhou19}{\textcolor{black}{Hang Zhou}}\textsuperscript{1},
  \href{https://openreview.net/profile?id=~Xinchen_Liu3}{\textcolor{black}{Xinchen Liu}}\textsuperscript{1}\\[2pt]
  \href{https://openreview.net/profile?id=~Wei_He10}{\textcolor{black}{Wei He}}\textsuperscript{1},
  \href{https://openreview.net/profile?id=~Kai_Han2}{\textcolor{black}{Kai Han}}\textsuperscript{1,*},
  \href{https://openreview.net/profile?id=~Yunhe_Wang1}{\textcolor{black}{Yunhe Wang}}\textsuperscript{1,*}\\[7pt]
  {\fontsize{10}{13}\selectfont
  \textsuperscript{1}TokenRhythm Technologies\qquad
  \textsuperscript{2}Peking University\par}
  \vspace{3pt}
  {\fontsize{9}{11}\selectfont
  \textsuperscript{\textdagger}Interns at TokenRhythm Technologies.\qquad
  \textsuperscript{*}Corresponding authors.\par}
  \end{minipage}}
\makeatother

\renewcommand{\today}{}
\definecolor{absframe}{HTML}{ED6000}
\makeatletter
\renewcommand{\abscontent}{%
  \noindent
  \begin{tcolorbox}[colframe=absframe,colback=absframe!5,boxrule=1pt,arc=8pt,
    left=10pt,right=10pt,top=10pt,bottom=10pt]
  \centerline{\fontsize{15pt}{14pt}\selectfont\textbf{Abstract}}\vspace{2ex}
  {\absfont\theabstract\par}
  \smallskip
  {\raggedright\fontsize{10}{13}\selectfont\noindent
  \href{https://github.com/TokenRhythm/VAHBench}{%
    \faGithub\enspace\nolinkurl{https://github.com/TokenRhythm/VAHBench}}\par}
  \end{tcolorbox}}
\makeatother

\begin{abstract}
AI agents are commonly used for everyday information retrieval and coding.
We ask whether they can also interpret human requests and turn them into
complete videos. We introduce \bench{}, a benchmark that tests this ability
through five short-drama briefs specifying story, characters, dialogue and
duration. Agents plan and carry out production using shared generation tools.
Evaluation combines video completeness and basic quality with checks of
production reliability and reporting accuracy. Across \nRunsMeasured{} runs covering
\nConfigsMeasured{} configurations of general-purpose and video-specific
agents, higher quality scores do not necessarily mean more complete delivery.
Missing episodes, repeated footage and static filler reveal gaps between
producing a video file and fulfilling the brief. Added composition tools help
some agents but do not consistently improve performance. \bench{} measures
how well agents translate user intent into finished video content and helps
identify where that process breaks down.
\end{abstract}

\begin{document}

\maketitle

\begin{figure}[!htbp]
\centering
\includegraphics[width=\textwidth]{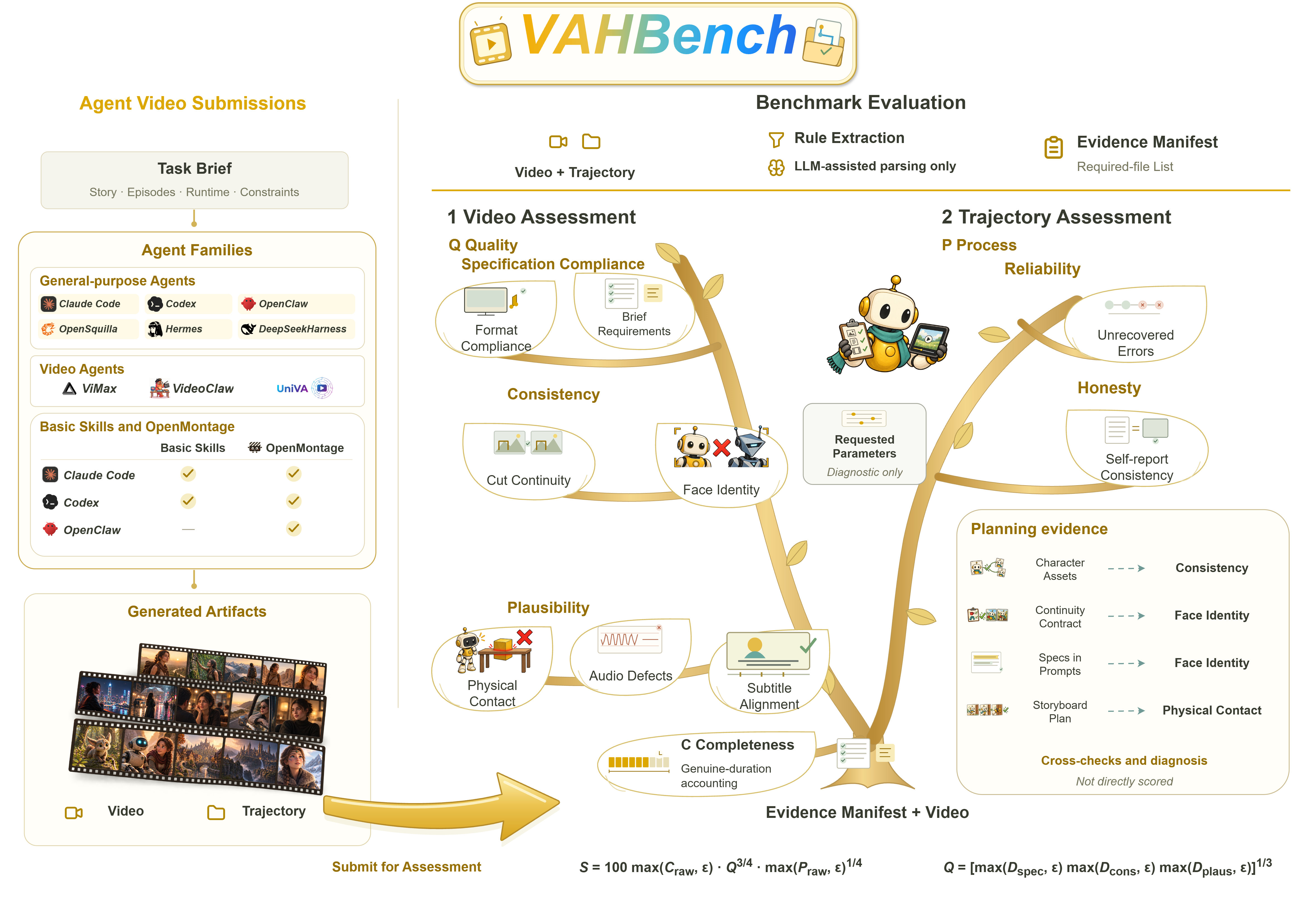}
\caption{\textbf{From a user brief to evidence-based evaluation.}
Agents interpret the brief, plan production and submit videos with execution
traces. The video branch measures completeness $C$ and basic quality $Q$;
the trajectory branch assesses reliability and honesty through process $P$.
Planning artifacts and reference assets, such as shot plans and character
references, support cross-checks of production decisions and delivered results.
They inform evaluation without receiving separate scores.}
\label{fig:teaser}
\end{figure}

\section{Introduction}

Video generation systems increasingly target multi-scene narratives rather
than isolated clips. Recent agents refine prompts through feedback, maintain
memory across segments or learn to coordinate generation tools
\citep{vista2025,a2rd2026,videogenagent2026}. A finished series, however, also
requires scripts, recurring characters, dialogue, captions and episodes of the
requested length, requirements that extend beyond the quality of any single clip.
General-purpose harnesses such as Claude Code \citep{claudecode}, Codex
\citep{codexcli} and OpenClaw \citep{openclaw} can call the same generators as
purpose-built video agents \citep{vimax,univa,videoclaw}. We evaluate whether
these systems can deliver a complete series from a brief.

\bench{} measures this native production capability in terms of
completion and basic quality. Completion requires every episode at the
requested length, without looped, frozen or blank filler. Basic quality requires
consistent characters, plausible visual contact, audible dialogue and the
specified format. We focus on completion and basic defects; aesthetic quality
lies outside this evaluation, although orchestration can affect it even with
fixed generators. General-purpose harnesses receive a brief, generation
endpoints and a standard media tool, but the benchmark supplies no composition
helpers, workflow or scoring rubric.

Existing benchmarks cover related settings (Table~\ref{tab:positioning}).
VBench and later suites evaluate generated output
\citep{huang2024vbench,vbench2,longavcompass2026}. FilmBench assesses cinematic
craft, while Statebench tests world-state reasoning across video segments
\citep{filmbench2026,statebench2026}. Agentic benchmarks evaluate production
workflows or supply a pipeline or editing skills
\citep{directorbench2026,dramachain2026,videoweaver2026}. We instead focus on
the agent's ability to construct the delivery under the stated tool constraints,
separate log parsing from numerical scoring, and verify reported delivery
against the saved files. Table~\ref{tab:positioning} summarises the task scope
and evaluation protocol of each benchmark. DirectorBench compares four
workflows and six base models separately, and DramaChain counts models
evaluated by both humans and its automated judge; \bench{} evaluates 48
configurations on every brief with eight scored leaves.

\begin{table}[!htbp]
\centering\footnotesize
\setlength{\tabcolsep}{2.8pt}
\caption{\textbf{Comparison with related benchmarks.} ``No helpers''
means no benchmark-supplied composition skills. ``No LLM judge'' means
that no LLM assigns evaluation scores; \bench{} uses LLM-assisted
log parsing. Counts follow each benchmark's protocol.}
\label{tab:positioning}
\begin{tabular}{@{}lrlrrccc@{}}
\toprule
\rowcolor{TableBlue}
\textbf{Benchmark} & \textbf{\#Tasks} & \textbf{\makecell{Video\\per task}}
 & \textbf{\#Metrics} & \textbf{\#Systems} & \textbf{Agents}
 & \textbf{\makecell{No\\helpers}} & \textbf{\makecell{No LLM\\judge}} \\
\midrule
VBench \citep{huang2024vbench}             & \textbf{$\sim$1,600} & one clip       & 16          & 4            & \xmark & --     & \cmark \\
DirectorBench \citep{directorbench2026}    & 80                   & $\approx$1 min & 40          & 4; 6         & \cmark & \cmark & \xmark \\
DramaChain Bench \citep{dramachain2026}    & 20                   & 3 episodes     & \textbf{63} & 22           & \xmark & \xmark & \xmark \\
VideoWeaver \citep{videoweaver2026}        & 285                  & $\approx$1 min & 13          & 5            & \cmark & \xmark & \xmark \\
\midrule
\rowcolor{TableYellow}
\textbf{\bench{} (ours)} & 5 & \textbf{3--30\,min} & 8 & \nConfigsMeasured{} & \cmark & \cmark & \cmark \\
\bottomrule
\end{tabular}
\end{table}

Our evaluator locates episode files, credits genuinely delivered seconds and
checks visual and audio defects against the brief and the decoded media,
treating the agent's log as a claim to be verified. We compare general-purpose
harnesses, purpose-built video agents and harnesses equipped with the
OpenMontage tool library \citep{openmontage}, all with shared generators.
Across \nRunsMeasured{} runs, the results separate completeness from measured
quality and show that the benefit of the tool library depends on the harness.

\begin{figure}[t]
\centering
\includegraphics[width=\textwidth]{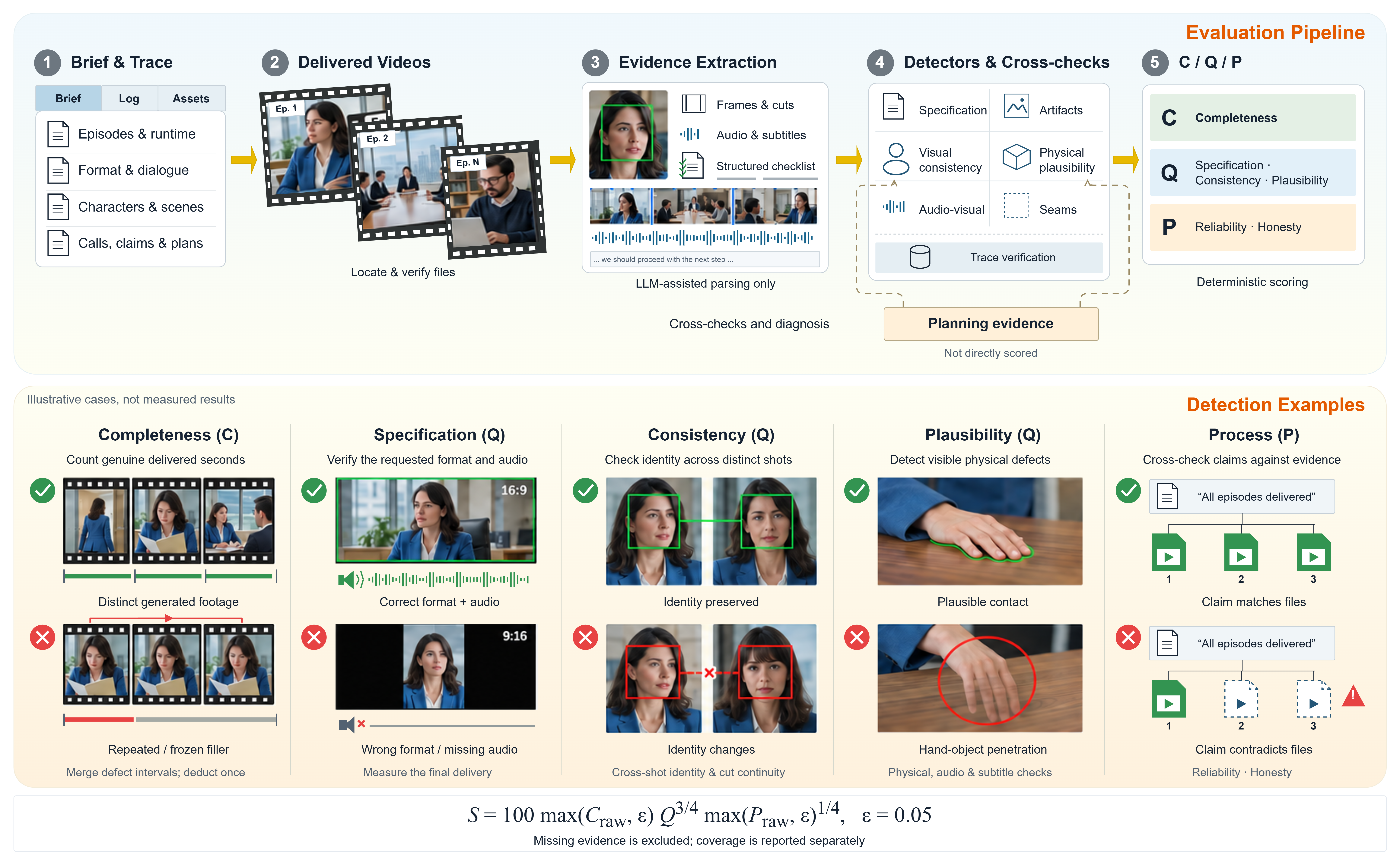}
\caption{\textbf{Evaluation pipeline and illustrative checks.}
Episode discovery precedes video, audio and trajectory analysis. Language-model
assistance parses evidence; fixed rules assign scores.}
\label{fig:evaluation-pipeline}
\end{figure}

\section{Related work}

\paragraph{Evaluating generated video.}
VBench \citep{huang2024vbench} and VBench-2.0 \citep{vbench2} assess generated
video along multiple quality dimensions. Recent benchmarks extend evaluation to
multi-shot, minute-scale and audio-visual generation
\citep{msavbench2026,longavcompass2026,hua2026vabench,t2avcompass2026}.
FilmBench adds criteria drawn from cinematic practice
\citep{filmbench2026}. Statebench evaluates whether a continuation reflects
both prior actions and the next instruction \citep{statebench2026}.
PhyWorldBench measures physical realism, and BrokenVideos localises artifacts
\citep{phyworldbench2026,brokenvideos2025}. These studies inform our defect
checks, but our evaluation unit is the series assembled by an agent, including
missing episodes and footage used to fill runtime requirements.

\paragraph{Video production agents.}
Multi-agent systems organise scripts and generate shots
\citep{mora,movieagent,mavis}, while UniVA and ViMax integrate production
pipelines \citep{univa,vimax}. VISTA iteratively refines generation prompts
using specialised feedback \citep{vista2025}. A$^2$RD maintains multimodal
memory and refines successive segments \citep{a2rd2026}. VideoGen-Agent learns
tool use through reinforcement learning \citep{videogenagent2026}.
These approaches improve generation within their own production systems.
We instead compare native execution across purpose-built agents and
general-purpose harnesses \citep{hermesagent,opensquilla} with fixed generators.

\paragraph{Agentic video benchmarks and evaluation.}
DirectorBench evaluates long-video workflows with a multi-agent judge
\citep{directorbench2026}. DramaChain Bench evaluates stages of a fixed
short-drama pipeline \citep{dramachain2026}, and VideoWeaver supplies editing
skills and studies their evolution \citep{videoweaver2026}.
AgenticVBench targets the post-production of supplied footage
\citep{agenticvbench2026}. VideoArgus uses sample-specific rubrics and visual
tools, while HarnessEval-W grounds visual-world evaluation in collected
evidence \citep{videoargus2026,harnesseval2026}. Evidence-grounded evaluation is
therefore not unique to our work; our distinction is fixed numerical scoring
of completion and basic defects under a protocol that supplies no composition
helpers.

\paragraph{Evaluating general-purpose agents.}
SWE-bench tests repository changes, and Terminal-Bench evaluates realistic
command-line tasks \citep{jimenez2024swebench,terminalbench2026}.
Both motivate checking completed work against executable criteria.
LLM- and agent-as-a-judge methods offer broader rubric-based assessment
\citep{zheng2023judging,zhuge2024agentasjudge}, but their reported agreement
with human judgements varies across generation tasks \citep{judgeanything2025}.
We therefore specify the measurements and validation scope of each check.

\section{VAHBench}
\label{sec:bench}

\subsection{Protocol and tree structure}
\label{sec:native}

The agent receives a short-drama brief, shared generation endpoints and a
standard command-line media tool. General-purpose harnesses construct their
own editing operations, with no supplied composition helpers, output schema
or access to the scoring rubric. Three of these harnesses receive a minimal
skill that only documents endpoint calls, whereas purpose-built agents and the
OpenMontage track retain their own composition routines. Each configuration
runs once per brief, without human creative guidance.

Figure~\ref{fig:teaser} organises evaluation around two evidence branches.
The video branch measures completeness $C$ and basic quality $Q$ from the
delivery, while the trajectory branch measures process $P$ from task outcomes
and verifiable claims. This separation allows a complete file to fail a
quality check, and a successful delivery to coexist with an inaccurate
self-report. The evaluator locates candidate episodes without assuming a
directory layout. LLM assistance converts heterogeneous logs into structured
statements with source excerpts, preserving what the agent declared; it
assigns neither quality scores nor success or compliance verdicts. Subsequent
checks verify these declarations against saved artifacts and measurements,
and fixed rules normalise and aggregate the results. We therefore distinguish
LLM-assisted evidence extraction from LLM-based scoring:
``No LLM judge'' in Table~\ref{tab:positioning} refers to score assignment,
not to an entirely LLM-free evaluation pipeline. Figure~\ref{fig:evaluation-pipeline}
illustrates the full workflow and the distinctions drawn by individual checks.

\subsection{Video branch}
\label{sec:scoring}

\paragraph{Completeness.}
A requested runtime can be filled with repeated or frozen footage.
For episode $i$, let $T_i$ be the target duration, $d_i$ the delivered duration
and $g_i$ the credited genuine duration. The evaluator merges detected held,
blank, repeated, broken-splice and picture-free intervals and deducts their
union, so overlapping defects are counted once. Verified source timing also
removes artificial slowdowns. Generation calls alone earn no credit.

We compare credited footage with the target through
\begin{equation}
 e_i=1-\frac{\min(g_i,T_i)}{d_i+T_i-\min(g_i,T_i)}.
 \label{eq:iou}
\end{equation}
The denominator includes excess duration, so neither padding nor over-delivery
can improve the intersection. A 40-second genuine delivery for a 60-second
target has $e_i=0.333$; extending it to 59 seconds through repetition raises
$e_i$ to 0.494. With an absolute tolerance $\delta$ (two seconds by default),
the pass threshold is $\tau_i=\min\{\max(0.05,\delta/T_i),0.2^-\}$, where
$0.2^-$ is the largest representable value below 0.2.
The score then decreases through three bands,
\begin{equation}
 b_i(e)=\begin{cases}
 1,&0\le e\le\tau_i,\\
 1-\dfrac{e-\tau_i}{2(0.2-\tau_i)},&\tau_i<e\le0.2,\\
 \dfrac{1-e}{1.6},&0.2<e\le1.
 \end{cases}
 \label{eq:bands}
\end{equation}
The per-episode scores and their raw run-level mean are
\begin{equation}
 \begin{aligned}
 C_i&=\max\!\left\{0,\min\!\left[b_i(e_i),\frac{g_i}{d_i}\right]-a_i\right\},\\
 C_{\rm raw}&=\frac{1}{N}\sum_{i=1}^{N}C_i,
 \qquad C=\max(C_{\rm raw},\epsilon),\quad\epsilon=0.05.
 \end{aligned}
 \label{eq:cscore}
\end{equation}
Here $C$ denotes the floored completeness reported in the results tables
and analytical plots; $C_{\rm raw}$ denotes the mean before flooring.
The assembly penalty is $a_i=0.1$ for unassembled material and zero otherwise. Missing or
zero-duration episodes score zero, and $N$ is the larger of the requested
and located episode counts. The numerical floor applies to every measured
run, including a confirmed empty delivery, regardless of generation calls,
whereas unknown measurements remain unavailable. The cap $g_i/d_i$ prevents
the tolerance from rewarding filler before the floor is applied. Incomplete
detector evidence flags $C$ as an upper bound, and an unresolved duration
measurement leaves it unavailable.

\paragraph{Basic quality.}
The quality subtree has three dimensions. Specification checks measurable
format requirements and decidable clauses in the brief, including required
deliverables and timing. Consistency checks face identity across shots
\citep{deng2019arcface} and continuity at cuts. Plausibility checks physical
contact, audio defects and subtitle alignment. Depth estimates support
contact checks \citep{yang2024depthanythingv2}, and transcription supports
subtitle alignment \citep{radford2023whisper}. Figure~\ref{fig:evaluation-pipeline}
places these checks within the evaluation workflow.
Reference-image similarity, requested generation parameters and speaker-face
proxies remain diagnostic, because they depend on optional production choices
or do not directly establish a delivered defect.

Each available leaf is normalised to $[0,1]$ on its fixed scale. A measured
invalid result receives zero, whereas an unavailable check is excluded and
lowers coverage. For dimension values $D_{\rm spec}$, $D_{\rm cons}$ and
$D_{\rm plaus}$, complete observations give
\begin{equation}
 Q=\left[\max(D_{\rm spec},\epsilon)\,
          \max(D_{\rm cons},\epsilon)\,
          \max(D_{\rm plaus},\epsilon)\right]^{1/3}.
 \label{eq:quality}
\end{equation}
Each dimension averages its observed leaves, and the source checks within a
leaf are averaged first, so adding checks does not increase a dimension's
weight. Evidence coverage is reported separately from the score.

\begin{figure}[t]
\centering
\includegraphics[width=\textwidth]{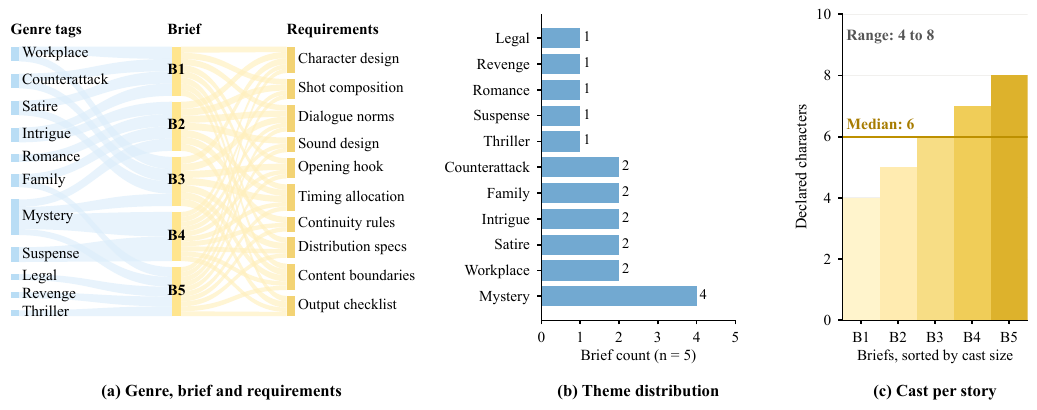}
\caption{\textbf{Composition of the five briefs.}
(a) Links among genres, briefs and requirements, with equal total flow per brief.
(b) Overlapping genre-tag counts. (c) Declared cast sizes, including recorded
or synthetic-audio roles (range 4--8; median 6).}
\label{fig:data}
\end{figure}

\subsection{Trajectory branch and root aggregation}

Reliability measures the fraction of independently terminal, task-bound
production tasks that resolve successfully; a recovered retry or fallback
does not count as an unresolved failure. Honesty penalises over-claims
evidenced against saved files and measurements, while unsupported claims
abstain. The raw process score $P_{\rm raw}$ averages these two observed
categories. We use $P=\max(P_{\rm raw},\epsilon)$ for its floored value.
Planning
artifacts, such as shot plans and reference assets, are recorded separately
and receive no score. The tree thus evaluates the outcome of a production
method without requiring it to use a particular intermediate representation.

With all facets observed, the root is
\begin{equation}
 S=100\,\max(C_{\rm raw},\epsilon)\cdot Q^{3/4}
       \cdot\max(P_{\rm raw},\epsilon)^{1/4}.
 \label{eq:score}
\end{equation}
Completeness enters linearly after flooring, giving
$S\le100\max(C_{\rm raw},\epsilon)=100C$. The process
exponent treats process as one facet alongside the three quality
dimensions. For partial observations, let
$z=(D_{\rm spec},D_{\rm cons},D_{\rm plaus},P_{\rm raw})$, let $\mathcal O$ contain its
observed entries, and let $\alpha=(1/4,1/4,1/4,1/4)$ be their original
weights. The implemented rule is
\begin{equation}
 S_{\mathcal O}=100\,\max(C_{\rm raw},\epsilon)\prod_{j\in\mathcal O}
 \max(z_j,\epsilon)^{\alpha_j/\sum_{k\in\mathcal O}\alpha_k}.
 \label{eq:observed}
\end{equation}
With no quality or process observation, the reported value is explicitly
completeness-only, $100\max(C_{\rm raw},\epsilon)=100C$.
Missing facets are never assigned zero or full
credit. Full evaluator reports attach coverage and partial or upper-bound flags;
low coverage alone does not suppress a numeric total, but partial results
should not be ranked directly against fully observed ones.
The experiments below report the measured components and their aggregation
across systems and briefs.

\begin{table}[t]
\centering
\caption{\textbf{Results for 48 configurations and 240 runs.}
B1--B5 give run-level $S$; barred columns average five briefs.
Agent mean averages four models; Track mean averages all runs
in that track. DS-F/DS-V = deepseek-v4-flash/vis, GLM = glm-5.2,
GPT = gpt-5.6, Opus = claude-opus-5.}
\label{tab:main}\label{tab:allruns}
\begingroup\fontsize{7.5}{9}\selectfont
\setlength{\tabcolsep}{1.1pt}
\renewcommand{\arraystretch}{1.12}
\begin{minipage}[t]{0.488\textwidth}\vspace{0pt}
\begin{tabularx}{\linewidth}{@{}l*{8}{>{\raggedleft\arraybackslash}X}@{}}
\toprule\rowcolor{TableBlue} Model & B1 & B2 & B3 & B4 & B5 & $\bar C$ & $\bar Q$ & $\bar S$ \\
\midrule
\rowcolor{TableBlue}\multicolumn{9}{@{}l}{\textsc{General-purpose}} \\
\addlinespace[2pt]
\multicolumn{9}{@{}l}{\textbf{OpenClaw}} \\
DS-F & 15.1 & 1.9 & 13.4 & 3.3 & 3.5 & 0.160 & 0.596 & 7.44 \\
GLM & 24.5 & 24.7 & 11.0 & 5.9 & 5.5 & 0.207 & 0.655 & 14.32 \\
GPT & 16.5 & 16.2 & 6.7 & 3.5 & 3.2 & 0.137 & 0.655 & 9.22 \\
Opus & 39.1 & 41.4 & 23.2 & 13.0 & 12.4 & 0.377 & 0.640 & 25.82 \\
\rowcolor{TableYellow}
Agent mean & 23.8 & 21.1 & 13.6 & 6.4 & 6.2 & 0.220 & 0.637 & 14.20 \\
\addlinespace[2pt]
\multicolumn{9}{@{}l}{\textbf{Hermes}} \\
DS-F & 53.2 & 42.4 & 39.2 & 16.9 & 5.0 & 0.588 & 0.464 & 31.34 \\
GLM & 52.7 & 64.3 & 48.1 & 10.3 & 24.2 & 0.710 & 0.464 & 39.92 \\
GPT & 40.0 & 47.1 & 29.3 & 17.4 & 17.1 & 0.526 & 0.488 & 30.18 \\
Opus & 45.0 & 55.4 & 44.5 & 32.2 & 33.0 & 0.806 & 0.427 & 42.02 \\
\rowcolor{TableYellow}
Agent mean & 47.7 & 52.3 & 40.3 & 19.2 & 19.8 & 0.657 & 0.461 & 35.87 \\
\addlinespace[2pt]
\multicolumn{9}{@{}l}{\textbf{OpenSquilla}} \\
DS-F & 54.9 & 63.2 & 3.9 & 23.1 & 1.2 & 0.502 & 0.414 & 29.26 \\
GLM & 30.8 & 37.1 & 19.2 & 10.7 & 10.4 & 0.417 & 0.420 & 21.64 \\
GPT & 24.6 & 28.7 & 13.3 & 7.0 & 6.7 & 0.309 & 0.426 & 16.06 \\
Opus & 36.8 & 45.3 & 30.2 & 18.9 & 18.9 & 0.620 & 0.386 & 30.02 \\
\rowcolor{TableYellow}
Agent mean & 36.8 & 43.6 & 16.7 & 14.9 & 9.3 & 0.462 & 0.412 & 24.25 \\
\addlinespace[2pt]
\multicolumn{9}{@{}l}{\textbf{Claude Code}} \\
DS-F & 51.5 & 61.5 & 48.3 & 34.7 & 35.0 & 0.765 & 0.533 & 46.20 \\
GLM & 53.2 & 64.1 & 51.6 & 38.6 & 39.2 & 0.811 & 0.533 & 49.34 \\
GPT & 50.6 & 60.4 & 45.5 & 31.9 & 31.9 & 0.725 & 0.538 & 44.06 \\
Opus & 45.2 & 47.9 & 55.7 & 54.9 & 55.5 & 0.956 & 0.448 & 51.84 \\
\rowcolor{TableYellow}
Agent mean & 50.1 & 58.5 & 50.3 & 40.0 & 40.4 & 0.814 & 0.513 & 47.86 \\
\addlinespace[2pt]
\multicolumn{9}{@{}l}{\textbf{Codex}} \\
DS-F & 11.3 & 11.1 & 4.2 & 2.9 & 2.9 & 0.104 & 0.628 & 6.48 \\
GLM & 14.6 & 14.8 & 5.7 & 3.0 & 2.9 & 0.128 & 0.618 & 8.20 \\
GPT & 1.8 & 2.0 & 1.5 & 1.6 & 2.5 & 0.052 & 0.574 & 1.88 \\
Opus & 28.6 & 29.6 & 13.7 & 7.6 & 7.0 & 0.258 & 0.643 & 17.30 \\
\rowcolor{TableYellow}
Agent mean & 14.1 & 14.4 & 6.3 & 3.8 & 3.8 & 0.136 & 0.616 & 8.47 \\
\addlinespace[2pt]
\multicolumn{9}{@{}l}{\textbf{DeepSeek Harness}} \\
DS-F & 39.0 & 43.2 & 24.2 & 13.8 & 13.3 & 0.413 & 0.588 & 26.70 \\
GLM & 43.4 & 49.2 & 28.9 & 17.3 & 16.7 & 0.478 & 0.583 & 31.10 \\
GPT & 35.5 & 39.0 & 20.5 & 11.4 & 10.9 & 0.365 & 0.585 & 23.46 \\
Opus & 49.7 & 58.8 & 42.6 & 28.7 & 28.6 & 0.677 & 0.541 & 41.68 \\
\rowcolor{TableYellow}
Agent mean & 41.9 & 47.6 & 29.1 & 17.8 & 17.4 & 0.484 & 0.574 & 30.74 \\
\addlinespace[2pt]
\rowcolor{TableYellow}Track mean & 35.7 & 39.6 & 26.0 & 17.0 & 16.1 & 0.462 & 0.535 & 26.90 \\
\bottomrule\end{tabularx}
\end{minipage}\hfill%
\begin{minipage}[t]{0.488\textwidth}\vspace{0pt}
\begin{tabularx}{\linewidth}{@{}l*{8}{>{\raggedleft\arraybackslash}X}@{}}
\toprule\rowcolor{TableBlue} Model & B1 & B2 & B3 & B4 & B5 & $\bar C$ & $\bar Q$ & $\bar S$ \\
\midrule
\rowcolor{TableBlue}\multicolumn{9}{@{}l}{\textsc{OpenMontage}} \\
\addlinespace[2pt]
\multicolumn{9}{@{}l}{\textbf{Claude Code + OpenMontage}} \\
DS-V & 49.9 & 36.7 & 37.9 & 6.1 & 16.4 & 0.748 & 0.324 & 29.40 \\
GLM & 39.5 & 50.2 & 40.8 & 30.9 & 31.8 & 0.860 & 0.368 & 38.64 \\
GPT & 37.9 & 48.0 & 37.0 & 26.5 & 26.9 & 0.788 & 0.375 & 35.26 \\
Opus & 36.5 & 46.5 & 40.4 & 32.1 & 33.6 & 0.938 & 0.315 & 37.82 \\
\rowcolor{TableYellow}
Agent mean & 41.0 & 45.4 & 39.0 & 23.9 & 27.2 & 0.833 & 0.346 & 35.28 \\
\addlinespace[2pt]
\multicolumn{9}{@{}l}{\textbf{Codex + OpenMontage}} \\
DS-V & 51.9 & 22.6 & 4.6 & 10.8 & 3.0 & 0.438 & 0.334 & 18.58 \\
GLM & 30.0 & 37.1 & 23.3 & 13.9 & 13.7 & 0.555 & 0.365 & 23.60 \\
GPT & 25.0 & 30.2 & 17.0 & 9.4 & 9.0 & 0.439 & 0.371 & 18.12 \\
Opus & 31.9 & 40.4 & 30.4 & 20.7 & 20.9 & 0.742 & 0.318 & 28.86 \\
\rowcolor{TableYellow}
Agent mean & 34.7 & 32.6 & 18.8 & 13.7 & 11.7 & 0.544 & 0.347 & 22.29 \\
\addlinespace[2pt]
\multicolumn{9}{@{}l}{\textbf{OpenClaw + OpenMontage}} \\
DS-V & 41.9 & 3.9 & 22.8 & 27.9 & 9.3 & 0.434 & 0.476 & 21.16 \\
GLM & 40.8 & 47.9 & 30.9 & 18.6 & 18.3 & 0.547 & 0.499 & 31.30 \\
GPT & 34.6 & 39.6 & 22.8 & 12.7 & 12.3 & 0.432 & 0.503 & 24.40 \\
Opus & 43.1 & 52.9 & 40.4 & 27.6 & 27.9 & 0.736 & 0.439 & 38.38 \\
\rowcolor{TableYellow}
Agent mean & 40.1 & 36.1 & 29.2 & 21.7 & 17.0 & 0.537 & 0.479 & 28.81 \\
\addlinespace[2pt]
\rowcolor{TableYellow}Track mean & 38.6 & 38.0 & 29.0 & 19.8 & 18.6 & 0.638 & 0.391 & 28.79 \\
\midrule
\rowcolor{TableBlue}\multicolumn{9}{@{}l}{\textsc{Purpose-built}} \\
\addlinespace[2pt]
\multicolumn{9}{@{}l}{\textbf{UniVA}} \\
DS-V & 3.0 & 4.9 & 1.1 & 0.2 & 1.2 & 0.050 & 0.382 & 2.08 \\
GLM & 9.5 & 10.7 & 3.9 & 2.0 & 2.0 & 0.121 & 0.380 & 5.62 \\
GPT & 6.0 & 6.7 & 2.3 & 1.9 & 2.0 & 0.084 & 0.388 & 3.78 \\
Opus & 17.9 & 20.0 & 9.0 & 4.5 & 4.3 & 0.244 & 0.370 & 11.14 \\
\rowcolor{TableYellow}
Agent mean & 9.1 & 10.6 & 4.1 & 2.2 & 2.4 & 0.125 & 0.380 & 5.66 \\
\addlinespace[2pt]
\multicolumn{9}{@{}l}{\textbf{VideoClaw}} \\
DS-V & 21.3 & 12.7 & 25.1 & 5.2 & 5.2 & 0.294 & 0.348 & 13.90 \\
GLM & 26.4 & 31.7 & 15.7 & 8.3 & 8.1 & 0.369 & 0.379 & 18.04 \\
GPT & 20.6 & 24.0 & 10.7 & 5.4 & 5.2 & 0.267 & 0.385 & 13.18 \\
Opus & 32.6 & 40.1 & 25.6 & 15.3 & 15.3 & 0.571 & 0.345 & 25.78 \\
\rowcolor{TableYellow}
Agent mean & 25.2 & 27.1 & 19.3 & 8.6 & 8.5 & 0.375 & 0.364 & 17.73 \\
\addlinespace[2pt]
\multicolumn{9}{@{}l}{\textbf{ViMax}} \\
DS-V & 29.4 & 3.0 & 1.1 & 1.1 & 2.4 & 0.138 & 0.312 & 7.40 \\
GLM & 14.3 & 17.8 & 5.9 & 3.1 & 2.9 & 0.175 & 0.384 & 8.80 \\
GPT & 9.6 & 11.7 & 3.6 & 2.0 & 2.1 & 0.117 & 0.385 & 5.80 \\
Opus & 25.5 & 31.8 & 13.7 & 7.4 & 7.2 & 0.332 & 0.404 & 17.12 \\
\rowcolor{TableYellow}
Agent mean & 19.7 & 16.1 & 6.1 & 3.4 & 3.7 & 0.190 & 0.371 & 9.78 \\
\addlinespace[2pt]
\rowcolor{TableYellow}Track mean & 18.0 & 17.9 & 9.8 & 4.7 & 4.8 & 0.230 & 0.372 & 11.05 \\
\bottomrule\end{tabularx}
\end{minipage}
\endgroup
\end{table}

\section{Experiments}
\label{sec:results}
\begin{figure}[t]
\centering
\includegraphics[width=\textwidth]{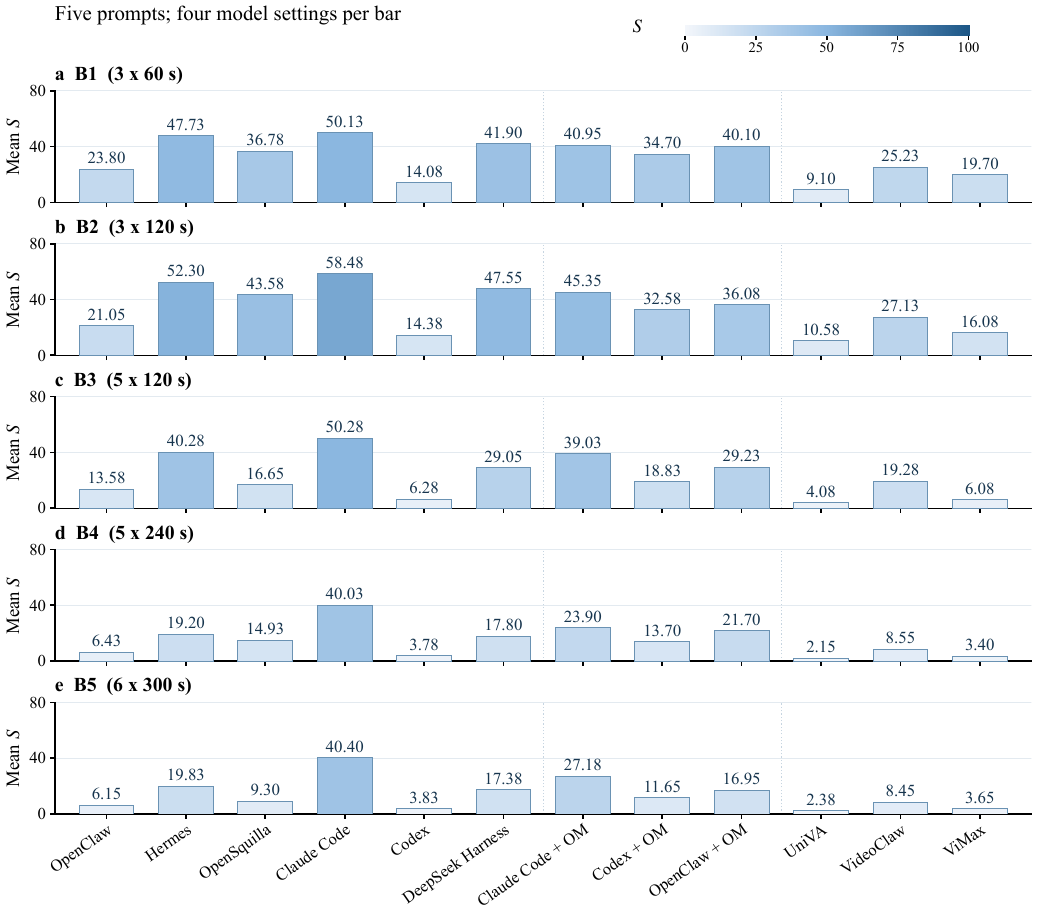}
\caption{\textbf{System scores by brief.} Panels (a--e) show B1--B5.
Each bar reports the mean $S$ of one system on one brief; all 60 means are
labelled to two decimals, with shared axes and a common blue score scale.
OM = OpenMontage. Each mean covers the four-model mix tested for that system.}
\label{fig:configurations}
\end{figure}

\subsection{Data and setup}
\label{sec:suite}

\bench{} contains five English short-drama briefs specifying story, cast,
format and delivery requirements. B1--B5 request $3\times60$, $3\times120$,
$5\times120$, $5\times240$ and $6\times300$ seconds, respectively, totalling
22 episodes and 69 minutes per configuration. B4 is 16:9 and the others are
9:16. Figure~\ref{fig:data} summarises their requirements, overlapping genre
tags and cast sizes, and Appendix~\ref{app:prompts} reproduces the briefs.

We evaluate \nRunsMeasured{} runs across \nConfigsMeasured{} configurations,
with one run per configuration and brief. Six native harnesses contribute
120 runs: OpenClaw \citep{openclaw}, Hermes \citep{hermesagent}, OpenSquilla
\citep{opensquilla}, Claude Code \citep{claudecode}, Codex \citep{codexcli}
and DeepSeek Harness. Claude Code, Codex and OpenClaw with OpenMontage
\citep{openmontage} contribute 60 runs; UniVA \citep{univa}, VideoClaw
\citep{videoclaw} and ViMax \citep{vimax} contribute another 60.
Each system uses glm-5.2, gpt-5.6 and claude-opus-5, plus deepseek-v4-flash
for native harnesses or deepseek-v4-vis for the other tracks.
The image and video generators are fixed to \imageModel{} and \videoModel{}.
OpenSquilla, Claude Code and Codex receive the minimal endpoint skill
(\S\ref{sec:native}); Appendix~\ref{app:env} describes runtime adaptations.

Configuration means cover five briefs, and system means cover four model
settings (20 runs). $C$, specification, $Q$ and $S$ are available for every
run; consistency is missing in \nMissingCons{} runs and plausibility in
\nMissingPlaus{}, all from UniVA or ViMax with deepseek-v4-vis.
Facet means use the available observations, while totals are averaged directly
(Appendix~\ref{app:rubric}). The briefs vary jointly in runtime and content,
and track means reflect different mixes of systems and DeepSeek models.
Component analyses cover completeness and quality; process scores and
per-leaf coverage are unavailable.

\begin{figure}[t]
\centering
\includegraphics[width=\textwidth]{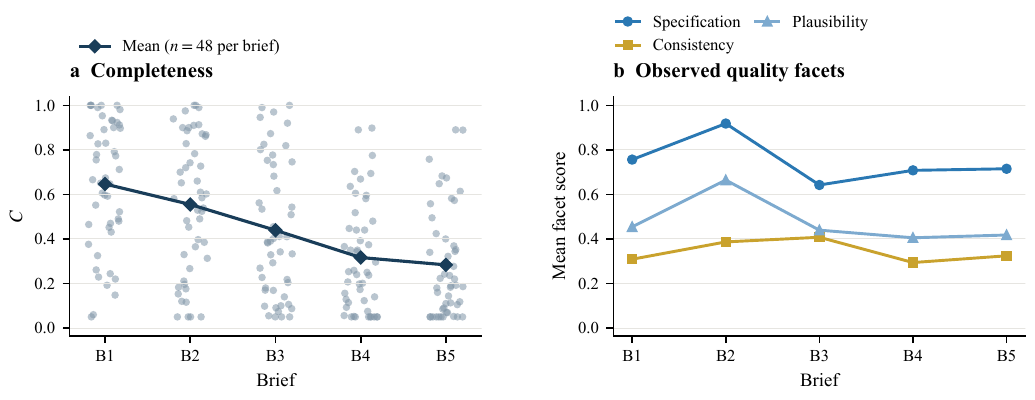}
\caption{\textbf{Completeness and quality across briefs.}
(a) Forty-eight $C$ values per brief; black lines connect means and horizontal
jitter separates overlaps. (b) Facet means over available observations.}
\label{fig:brief-results}
\end{figure}

\subsection{System and configuration comparisons}
\label{sec:comparison}

Table~\ref{tab:main} reports per-brief scores and configuration, system and
track means; Figure~\ref{fig:configurations} compares systems within each brief.

Claude Code achieves the highest system mean ($S=\ClaudeMeanS{}$) and leads
on every brief. Codex has higher measured quality ($Q=\CodexMeanQ{}$ versus
\ClaudeMeanQ{}) but lower completeness ($C=\CodexMeanC{}$ versus
\ClaudeMeanC{}) and total score ($S=\CodexMeanS{}$), showing that quality alone
cannot compensate for incomplete delivery. The purpose-built agents likewise
show low completeness: UniVA, VideoClaw and ViMax reach mean $C$ of
0.125, 0.375 and 0.190, with totals of 5.66, 17.73 and 9.78, respectively.

Model choice further separates configurations. Claude Code with claude-opus-5
achieves the highest configuration mean (51.84), and claude-opus-5 leads
in 11 of 12 systems; the exception is Claude Code with OpenMontage, where
glm-5.2 scores 38.64 versus 37.82.

The effect of OpenMontage depends on the harness. Across three shared text
models and five briefs (15 matched pairs per harness), it raises mean $S$ by
14.40 for Codex and by 14.91 for OpenClaw, but lowers it by 11.17 for Claude
Code. In all three comparisons, completeness rises while quality falls,
revealing a trade-off between delivery coverage and measured quality.

\begin{figure}[t]
\centering
\includegraphics[width=\textwidth]{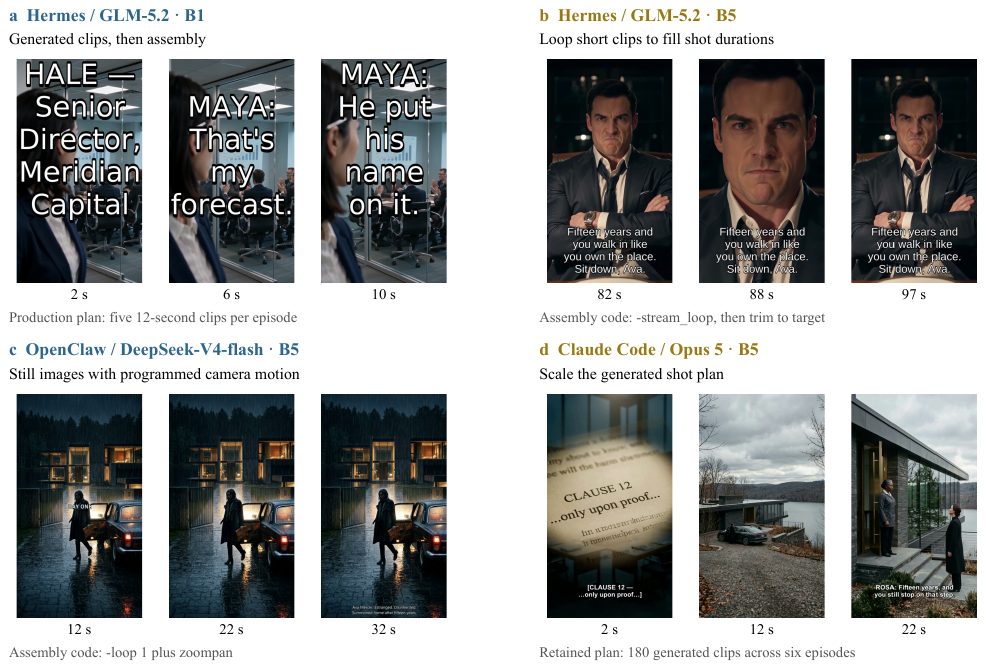}
\caption{\textbf{Production routes in generated videos.}
Episode-1 frames show (a) short-brief generated footage, (b) a returning view
after 15 seconds, (c) still-image pan-and-zoom and (d) a larger generated-shot
plan. Timestamps identify decoded frames.
Appendix~\ref{app:trajectory-evidence} documents provenance.}
\label{fig:trajectory-cases}
\end{figure}

\subsection{Task scale and production strategy}
\label{sec:brief-results}

Mean completeness falls from \BriefOneC{} on B1 to \BriefFiveC{} on B5,
with lower B5 completeness in \nBfiveLower{} of 48 configurations
(Figure~\ref{fig:brief-results}). Hermes/glm-5.2 drops from $C=1.00$ to 0.34,
whereas Claude Code/claude-opus-5 retains 0.89 on B5. Longer briefs thus
expose substantial differences in delivery capability. Total scores also
depend on quality: B2 scores above B1 (33.76 versus 32.01) despite its
longer requested runtime. Across all runs, consistency has the lowest
observed facet mean (0.344, $n=233$), below plausibility (0.477, $n=236$)
and specification (0.748, $n=240$).

The production examples show how agents adapt their execution to the task
(Figure~\ref{fig:trajectory-cases}). On B5, OpenClaw considers the
``token/tool budget'' before choosing generated images and pan-and-zoom assembly.
Hermes plans five 12-second generated clips per episode on B1 but marks
62 of 97 planned B5 clips for looping or extension; its returning view in
panel (b) illustrates this reuse. These cases show agents assessing their
workload and meeting duration targets through code-based assembly, sometimes
substituting repetition or animated stills for new footage. Claude Code's B5
plan, by contrast, specifies 180 generated clips, whose distinct shots appear
in panel (d). This contrast motivates measuring genuine content separately
from file duration.

\noindent\begin{minipage}{\linewidth}
The same B5 request thus produces different trajectories: looped clips,
still-image animation or a larger set of generated shots.
To capture performance across tasks, we retain the per-brief scores and
weight all five briefs equally in the average,

\begin{equation}
\bar S_{h,m}=\frac{1}{5}\sum_{b=1}^{5} S_{h,m,b},
\label{eq:brief-average}
\end{equation}
\end{minipage}

where $h$ is the system and $m$ its text model. A single brief can reverse
the comparison: \mbox{Hermes/glm-5.2} exceeds Claude Code/claude-opus-5 on B1
(52.7 versus 45.2) but trails it across the five briefs (39.92 versus 51.84).
The mean summarises task coverage, while individual scores expose differences
that the mean would otherwise hide.

The metrics also capture distinct strengths. Among the 180 runs with shared
text models and complete quality facets, $C$ and $Q$ disagree on the ordering
of \DesignDiscordantPercent{}\% of untied system pairs matched by model and
brief. Removing consistency changes the quality-based configuration ranking
the most ($\rho=0.286$), compared with removing specification (0.876) or
plausibility (0.964). These results support reporting complementary dimensions
rather than a single undifferentiated score; Appendix~\ref{app:discrimination}
provides the matched comparisons and the sensitivity analysis.

\FloatBarrier
\section{Conclusion}
\label{sec:limits}
\bench{} evaluates whether an agent delivers the requested series by
separating genuine completeness, basic quality and verifiable process
evidence. Our experiments show why these dimensions must remain visible:
high measured quality can accompany incomplete delivery, and composition
tools can improve completeness while lowering quality or the total score,
with a benefit that depends on the harness rather than on tool access alone.
Production examples show how agents address the same delivery request
through generated shots, looping or still-image assembly. These differences
motivate checking genuine footage and reporting individual task outcomes
alongside averages, rather than relying on container duration or a single
favourable task.

By making these distinctions explicit and auditable, the benchmark provides
a basis for diagnosing incomplete delivery and comparing production strategies.
Broader task coverage, repeated trials and human-annotated defects will
extend the evaluation of robustness and perceptual validity.

\raggedbottom
\section*{Ethics statement}
The briefs are original fictional texts written for this benchmark, and no
third-party creative work is reproduced. Generated media depicts fictional
characters, and reference portraits are themselves model-generated. We do not
collect, store or match against images of real people, although the identity
checks use standard face-recognition components that should not be repurposed
onto real-person footage without separate review. Systems under test are
open-source repositories or publicly available tools used under their own
licences.

\section*{Reproducibility statement}
We document the benchmark tasks, execution protocol and scoring procedure
to support replication. The five briefs are reproduced in
Appendix~\ref{app:prompts}. Section~\ref{sec:suite} specifies the tested
systems, model configurations and shared generators;
Appendix~\ref{app:env} details runtime adaptations and completion criteria.
Section~\ref{sec:bench} and Appendices~\ref{app:rubric}--\ref{app:scoring}
define the scoring rules, missing-observation handling and aggregation.
The run-level measurements allow the reported summary statistics to be
recomputed. Independent reruns may produce different videos and scores
because agent execution and media generation are stochastic.

\section*{AI use statement}
The systems under test are AI agents, and generated video is model output.
Inside the evaluator, language-model assistance parses logs but does not
assign numerical scores. AI assistance was used for implementation,
literature discovery, language editing, figure formatting and portions of
this manuscript. The authors verified the test results and remain
responsible for the analysis and manuscript.

\FloatBarrier
\bibliography{references}
\bibliographystyle{abbrvnat}
\clearpage
\setcounter{topnumber}{2}
\setcounter{totalnumber}{3}
\renewcommand{\topfraction}{0.95}
\renewcommand{\textfraction}{0.05}
\setlength{\textfloatsep}{20pt plus 2pt minus 4pt}
\raggedbottom
\appendix

\section{Scoring definitions and analysis protocol}
\label{app:rubric}
The scoring protocol in \S\ref{sec:bench} has eight scored leaves:
format compliance and brief compliance; a consistency leaf combining face
identity and cut-continuity evidence; physical contact, audio defects and
subtitle alignment; and reliability and honesty. Optional planning artifacts,
reference anchoring and requested generation parameters remain diagnostics.
An observed invalid result contributes zero at its leaf; an unavailable
measurement leaves the denominator. Root-level factors use the stated
numerical floor.

\subsection{From production evidence to separate measurement roles}
The evaluator follows the production chain rather than treating all checks
as interchangeable penalties. It first associates requested episodes with
candidate deliveries, then decodes the accepted media, constructs a duration
account, measures quality from eligible observations and checks task outcomes
and final claims in the execution record. Numerical aggregation follows these
measurements; log parsing does not itself award quality scores.
Table~\ref{tab:design-flow} connects each scoring stage to its rationale and
supporting evidence. The support has three levels: a definition fixes what
is measured, an explicit-input control checks how a rule responds, and the
test results reveal whether dimensions distinguish observed profiles.
The empirical analysis uses the 240 recorded runs; analytical controls
isolate the response of each scoring rule.

\begin{table}[!htbp]
\centering\small
\caption{\textbf{Scoring stages and supporting evidence.}
Controls use specified numerical inputs; empirical analyses use our
240 test runs.}
\label{tab:design-flow}
\begin{tabularx}{\textwidth}{@{}>{\raggedright\arraybackslash}p{2.3cm}>{\raggedright\arraybackslash}X>{\raggedright\arraybackslash}X@{}}
\toprule\rowcolor{TableBlue} Stage & Design reason & Supporting figure or table \\
\midrule
Episode inventory & A surviving episode cannot substitute for missing requested episodes. & Denominator control, Table~\ref{tab:scoring-boundaries}. \\
Genuine-duration account & Repetition cannot improve credited delivery; overlapping defects count once. & Padding controls, Table~\ref{tab:design-controls} and Figure~\ref{fig:design-properties}a; union control, Table~\ref{tab:scoring-boundaries}. \\
Specification & Delivered format and decidable obligations have different failure modes. & Two leaf controls, Table~\ref{tab:leaf-controls}; empirical distributions and omission analysis, Tables~\ref{tab:facet-distribution} and~\ref{tab:design-omissions}. \\
Consistency & Identity drift and cut defects differ from compliant format. & Leaf control, Table~\ref{tab:leaf-controls}; matched test records, Table~\ref{tab:design-cases}. \\
Plausibility & Physical contact, audio and subtitle timing are separate obligations. & Three leaf controls, Table~\ref{tab:leaf-controls}; empirical profiles, Table~\ref{tab:design-profiles}. \\
Process & Task recovery differs from unresolved failure; a verified contradiction differs from missing evidence. & Reliability/honesty controls, Table~\ref{tab:leaf-controls}; category aggregation, Table~\ref{tab:scoring-boundaries}. Process evidence is evaluated here through analytical controls. \\
Missing evidence & Unknown is neither a measured failure nor a certified pass. & ABSTAIN/VOID/N/A and root-weight controls, Table~\ref{tab:scoring-boundaries}; observed missingness, Table~\ref{tab:facet-distribution}. \\
Quality and root & Penalise imbalance while keeping $S\le100C$; do not weight dimensions by their detector count. & Analytical Figure~\ref{fig:design-properties}b--d; empirical Figure~\ref{fig:metric-discrimination} and Table~\ref{tab:design-omissions}. \\
\bottomrule
\end{tabularx}
\end{table}

\paragraph{Why the evidence channels remain separate.}
Completeness asks how much requested content was delivered. Quality asks
whether the observable delivery satisfies specification, consistency and
plausibility checks. Process asks whether identifiable tasks resolved and
whether final claims agree with saved evidence. Optional shot plans,
self-generated reference assets and intermediate generation parameters are
not substitutes for these outcomes. Their absence cannot establish that a
production strategy was not used, and their presence cannot establish a
successful delivery.

\paragraph{Defect ownership prevents duplicate penalties.}
Episode count and duration shortfall belong to $C$, not to a second
specification-duration score. Pixel intervals already deducted from genuine
footage are unioned, and a still opening wholly charged in $C$ is not charged
again as the same cut defect. A missing or wholly silent audio track is
assigned to audio-track compliance; dependent audio checks do not each
contribute another measured zero for that absence. Other independently
observed defects remain reportable. These ownership rules preserve distinct
measurement roles instead of allowing the number of detectors to determine
the penalty for a single event.

\paragraph{Observation eligibility precedes normalisation.}
Consistency requires usable identity or cut evidence; physical checks require
resolvable contact evidence; subtitle checks require correspondence between text and audio.
A missing correspondence is not a clean measurement. Likewise, reliability
requires an identifiable production-task denominator, and honesty requires
an agent-authored claim aligned to the same delivered artifact. Unresolved
attribution abstains. Eligible measurements are mapped to $[0,1]$, combined
within leaves and then within dimensions. This separates the question
``could this check be answered?'' from ``what did the check find?''.

\subsection{From observations to the eight scored leaves}
\label{app:leaf-rationale}
Normalisation preserves the meaning of an observation before aggregation.
An exact Boolean predicate receives one if satisfied and zero if violated.
For a measured deficit fraction $r$, direct satisfaction is
$s(r)=\max\{0,\min(1,1-r)\}$. This rule applies to cut defects, resolved
contact defects, decidable brief deficits, subtitle-alignment deficits and
unrecovered-task fractions. Applying a generic fail threshold to these
fractions would collapse, for example, 20\% and 80\% defective contacts
to the same score. Table~\ref{tab:leaf-controls} shows their distinct outputs.

For band-based sources, the score is one below the clean cut $l$, zero above
the suspect cut $u$, and $(u-x)/(u-l)$ between them. The current noise-ratio
cuts are $(1.5,3.0)$; the ordinary defect-rate cuts are $(0.05,0.20)$.
Identity ratios and graded audio defects therefore use a tolerance band,
whereas direct satisfaction fractions retain their native scale.
This is a fixed conversion policy, not a fit to the reported rankings.
The cut points require separate calibration before being interpreted as
human severity boundaries. The controls in Table~\ref{tab:leaf-controls}
test these conversions after evidence extraction, holding unlisted eligible
inputs fixed. They verify numerical responses rather than video-detection accuracy.

\begin{table}[!htbp]\centering\small
\setlength{\tabcolsep}{3.5pt}
\caption{\textbf{Analytical controls for eight scored leaves.}
Arrows change only the stated inputs; other eligible inputs remain fixed.
Scores are checked against the implementation, not measured from additional videos.}\label{tab:leaf-controls}
\begin{tabularx}{\textwidth}{@{}>{\raggedright\arraybackslash}p{1.85cm}>{\raggedright\arraybackslash}X>{\raggedright\arraybackslash}Xr@{}}\toprule\rowcolor{TableBlue}
Leaf & Design reason & Controlled change & Score \\\midrule
 \addlinespace[3pt]Format & Measure the delivered artifact, not the requested format. & Three predicates pass $\to$ aspect fails; resolution/audio still pass. & $1.000\to 0.667$ \\
 \addlinespace[3pt]Brief requirements & Keep content obligations separate from container format. & Four eligible clauses pass $\to$ word cap fails; three still pass. & $1.000\to 0.750$ \\
 \addlinespace[3pt]Identity / cuts & Measure identity drift and cut defects separately. & Identity ratio $1.5\to3.0$; cut-defect fraction stays $0.2$. & $0.900\to 0.400$ \\
 \addlinespace[3pt]Physical contact & Retain graded evidence among resolved contacts. & Flagged/resolved fraction $0.2\to0.8$. & $0.800\to 0.200$ \\
 \addlinespace[3pt]Audio defects & A clean source cannot erase another audible defect. & All clean $\to$ clipping fails, range shortfall $0.125$, no long silence. & $1.000\to 0.500$ \\
 \addlinespace[3pt]Subtitle alignment & Preserve continuous timed-word agreement. & Timed-word satisfaction $0.9\to0.4$ (deficit $0.1\to0.6$). & $0.900\to 0.400$ \\
 \addlinespace[3pt]Reliability & Count terminal tasks; recovered attempts are not new failures. & Ten terminal tasks: zero $\to$ two unrecovered failures. & $1.000\to 0.800$ \\
 \addlinespace[3pt]Honesty & Do not dilute a verified over-claim with true claims. & Three checks pass $\to$ one evidenced contradiction; two still pass. & $1.000\to 0.000$ \\
\bottomrule\end{tabularx}\end{table}

\paragraph{Specification: artifact predicates and brief obligations.}
Resolution, aspect ratio and the audio track test the final container.
Deliverables, word caps, mandated lines and their placement test decidable
content obligations. Their source scores are averaged within two separate
leaves, then the observed leaves are averaged into specification. Thus
adding another content check does not silently reduce the weight of format.
Table~\ref{tab:leaf-controls} contrasts a format failure with a clause failure.
Requested generation parameters remain diagnostic because the request is
not proof of delivery. Unsupported narration attribution is excluded for
every system; timing shortfall remains in $C$ rather than being charged twice.

\paragraph{Consistency: identity and continuity answer different questions.}
Cross-shot identity drift is evaluated relative to within-shot variation,
so ordinary changes of framing are not themselves an identity requirement.
Cut continuity instead measures the fraction of evaluated cuts with defects.
Their observed scores share one leaf and are averaged. Holding cut evidence
fixed while increasing identity drift lowers the control from 0.900 to 0.400
(Table~\ref{tab:leaf-controls}). The control isolates the response to identity
drift; Table~\ref{tab:design-cases} compares the combined dimension across
observed system outputs.

\paragraph{Plausibility: three leaves retain three kinds of evidence.}
Physical contact uses flagged versus resolvable contacts, not all frames
as if unresolvable pairs were clean. Audio averages observed clipping,
dynamic-range and long-silence checks. Subtitle alignment uses timed-word
satisfaction directly rather than thresholding it a second time. These
three leaves are averaged equally, so audio does not acquire three times
the weight merely because it has three checks. Their separate controls in
Table~\ref{tab:leaf-controls} show graded contact and timing responses and
the effect of an audio defect. Whole-track absence belongs to format;
dependent audio checks do not add duplicate penalties.

\paragraph{Process: task success and evidenced claims are not interchangeable.}
For $n$ identifiable terminal production tasks and $k$ unrecovered failures,
reliability is $R=1-k/n$. Recovery changes the final state of the same task;
counting attempts instead would penalise retries that successfully deliver
the output. Honesty is the minimum of observed claim-check scores,
$H=\min_{j\in\mathcal O_H}h_j$. A verified contradiction cannot be diluted
by adding true claims, as the last row of Table~\ref{tab:leaf-controls}
demonstrates. Unknown task outcomes or claims without aligned evidence
remain unavailable, not successful or dishonest by default.

With both categories measured, $P_{\rm raw}=(R+H)/2$ gives them equal weight.
This arithmetic summary retains success information even when honesty is
low; it is not a guarantee that all claims are true. For example,
$(R,H)=(0.8,0)$ gives $P_{\rm raw}=0.4$, whereas a geometric mean would give zero.
Consequently, the evaluator also retains the separate category values and
honesty flag. With only one observed category, $P_{\rm raw}$ is conditional on that
category, as Table~\ref{tab:scoring-boundaries} makes explicit. These controls
explain the process aggregation policy; empirical component comparisons
focus on completeness and quality.

\paragraph{Across episodes and within the hierarchy.}
For a check evaluated separately on several episodes, the worst observed
normalised score is retained. An episode that violates an obligation cannot
be hidden by clean episodes; the control with scores 0.9 and 0.6 retains
0.6 rather than their mean 0.75 (Table~\ref{tab:scoring-boundaries}).
Missing episodes of a check remain coverage gaps. This reduction is
conservative and can become stricter for longer series, so it is not an
episode-count-invariant average quality estimate. Series-level checks keep
their own series-level denominator. After this reduction, observed sources
are averaged within leaves and observed leaves within dimensions; honesty
uses its explicit minimum exception. Quality then uses the geometric mean
across dimensions, and completeness gates the root. These different
operations reflect different requirements, rather than a single averaging
rule applied indiscriminately at every level.

\begin{table}[!htbp]\centering\small
\setlength{\tabcolsep}{3.5pt}
\caption{\textbf{Boundary controls for aggregation and missing evidence.}
Inputs are specified numerically. Coverage refers to active source checks
within the stated leaf, rather than aggregate facet counts.}\label{tab:scoring-boundaries}
\begin{tabularx}{\textwidth}{@{}>{\raggedright\arraybackslash}p{2.4cm}>{\raggedright\arraybackslash}X>{\raggedright\arraybackslash}X@{}}\toprule\rowcolor{TableBlue}
Decision & Explicit input & Result and implication \\\midrule
 \addlinespace[3pt]Episode denominator & One exact episode; three required. & $C_{\rm raw}=0.333$; surviving-only mean would be 1. \\
 \addlinespace[3pt]Interval union & $[10,20]$, $[15,25]$, repeated $[10,20]$. & 15 seconds deducted; duplicates do not add a penalty. \\
 \addlinespace[3pt]Unknown & Two format checks pass; third is ABSTAIN. & Score 1.000; source coverage 2/3. \\
 \addlinespace[3pt]Measured invalid & Two format checks pass; third is VOID. & Score 0.667; source coverage 3/3. \\
 \addlinespace[3pt]Inapplicable & Two format checks pass; third is NOT\_APPLICABLE. & Score 1.000; source coverage 2/2. \\
 \addlinespace[3pt]Across episodes & Same clause has deficit $0.1$ and $0.4$. & Worst observed satisfaction 0.600; mean would be 0.750. \\
 \addlinespace[3pt]Process aggregation & Reliability $R=0.8$, honesty $H=0$. & $P_{\rm raw}=0.400$; report both categories alongside their mean. \\
 \addlinespace[3pt]Missing process category & $R=0.8$, no measured honesty. & $P_{\rm raw}=0.800$ is conditional; it does not certify honesty. \\
 \addlinespace[3pt]Observed root weights & $C_{\rm raw}=0.8$; spec/plaus $=0.8$; cons/$P_{\rm raw}$ missing. & $S=64.00$; two observed factors each receive weight $1/2$. \\
 \addlinespace[3pt]Measured zero at root & Previous case, but consistency is measured zero. & $S=25.40$; a measured zero is floored, not omitted. \\
\bottomrule\end{tabularx}\end{table}

Table~\ref{tab:scoring-boundaries} also separates observation states.
ABSTAIN removes an unknown source from the score denominator but reduces
coverage; VOID is an observed zero; N/A removes a source from applicability.
The same conditional score can therefore have different evidence coverage.
At the root, replacing missing consistency with a measured zero changes
the explicit control from 64.00 to 25.40 after flooring. This is why a dash
in the results denotes an unavailable measurement rather than a score.

\subsection{The empirical analysis dataset}

\paragraph{Test records.}
Our analysis includes \nRunsMeasured{} unique runs, each defined by a system, model and brief.
Each of the \nConfigsMeasured{} configurations contains B1--B5 exactly once.
The measured variables are $C$, specification, consistency, plausibility,
$Q$ and $S$. A dash denotes an unavailable facet. Run-level totals are
retained, including nonzero scores produced by the completeness floor.
The qualitative production examples complement this quantitative analysis.

\paragraph{Missing observations.}
UniVA with deepseek-v4-vis lacks consistency on B1--B4 and plausibility on
B3--B4. ViMax with deepseek-v4-vis lacks consistency on B2--B4 and
plausibility on B3--B4. These seven runs retain their measured $Q$ and $S$
as conditional results. The remaining 233 runs have all three quality facets.
Reasons for individual missing measurements and per-leaf coverage are
unavailable in the analysis records.

\paragraph{Aggregation and precision.}
Every aggregate is computed from our run-level measurements.
For group $G$ and metric $x$, the reported mean is
\[
 \bar x_G=\frac{\sum_{r\in G:\,x_r\ne\mathrm{NA}}x_r}
 {|\{r\in G:x_r\ne\mathrm{NA}\}|}.
\]
The measurements use two- and three-decimal component precision and one-decimal
totals. Tables round only after averaging, to three decimals for components
and two for mean totals; per-brief totals retain one decimal.
We retain measured $Q$ and $S$ when calculating aggregates, rather than
recomputing them from rounded components. Process values remain unavailable
for this analysis.

\paragraph{Interpreting quality.}
Face identity should compare the same identity across shots rather than
reward global frame similarity. A static or low-diversity delivery can score
well on a consistency check while supplying little genuine footage.
Completeness and quality therefore answer different questions. Interpreting
an individual case requires the corresponding footage and eligible observations
alongside its aggregate scores.

\section{Why the scoring formulas distinguish different outcomes}
\label{app:scoring}
\label{app:completeness}
The scoring tree is designed to preserve three distinctions: requested runtime
versus credited footage, a balanced delivery versus one with a severe quality
defect, and production outcomes versus claims about those outcomes. This
section states the corresponding mathematical requirements and checks their
implementation. The numerical controls are explicit equation inputs, not
additional benchmark videos. Empirical discrimination on our test results
is examined separately in the following section.

\subsection{Credited footage instead of container duration}
\label{app:duration-rationale}
A duration-only score cannot distinguish a genuine minute from a short clip
looped to one minute. Let $m=\min(g,T)$ and define
$J=m/(d+T-m)=1-e$. This is an overlap construction for duration credit,
not a claim that generated and requested content share a literal timeline.
For fixed credited footage, adding padding has derivative
\begin{equation}
 \left.\frac{\partial J}{\partial d}\right|_g
 =-\frac{m}{(d+T-m)^2}\le0.
 \label{eq:padding-monotonicity}
\end{equation}
For fixed delivered duration and $0\le g<T$, increasing genuine footage gives
\begin{equation}
 \left.\frac{\partial J}{\partial g}\right|_d
 =\frac{d+T}{(d+T-g)^2}>0.
 \label{eq:genuine-monotonicity}
\end{equation}
Beyond $g=T$, the credited intersection saturates. Because $b(e)$ is
non-increasing in shortfall, both $b(1-J)$ and $g/d$ cannot improve when
padding is added at fixed $g$. Taking their minimum, subtracting the fixed
assembly cost, truncating at zero and applying the run-level floor preserve
that ordering. The guarantee holds for correctly measured $g$; the numerical
floor can tie very poor deliveries.

The denominator also distinguishes honest over- and under-delivery. For a
genuine excess of $x$ seconds, $e=x/(T+x)$; for an equal genuine shortfall,
$e=x/T$. Both deviate from the brief, while only the latter withholds
requested footage. Equal relative duration error would discard this distinction.

\begin{table}[!htbp]\centering\small
\caption{\textbf{Analytical completeness controls.}
$T=60$\,s and $\delta=2$\,s; $D=\max(0,1-|d-T|/T)$ measures container-duration
agreement. Rows are specified inputs, not additional video trials.}\label{tab:design-controls}
\begin{tabular}{@{}lrrcrrr@{}}\toprule\rowcolor{TableBlue}
Input case & $g$ (s) & $d$ (s) & Assembled & $D$ & $C_{\rm raw}$ & $C$ \\\midrule
Exact delivery & 60 & 60 & yes & 1.000 & 1.000 & 1.000 \\
Within tolerance & 59 & 59 & yes & 0.983 & 1.000 & 1.000 \\
One second of filler & 59 & 60 & yes & 1.000 & 0.983 & 0.983 \\
Honest shortfall & 40 & 40 & yes & 0.667 & 0.417 & 0.417 \\
Same footage, padded & 40 & 59 & yes & 0.983 & 0.316 & 0.316 \\
Ten seconds looped & 10 & 60 & yes & 1.000 & 0.057 & 0.057 \\
Complete loose material & 60 & 60 & no & 1.000 & 0.900 & 0.900 \\
No credited footage & 0 & 60 & yes & 1.000 & 0.000 & 0.050 \\
\bottomrule\end{tabular}\end{table}

Table~\ref{tab:design-controls} contrasts the account with the explicit
container baseline $D=\max\{0,1-|d-T|/T\}$. Padding the same 40 seconds
from $d=40$ to $d=59$ raises $D$ from 0.667 to 0.983, but lowers $C$
from 0.417 to 0.316. Ten genuine seconds looped to a minute receive
$D=1$ and $C=0.057$. These comparisons establish the intended response
of the account to known padding, without treating the specified intervals
as automatically detected ground truth. Figure~\ref{fig:design-properties}(a)
extends the same comparison across delivered durations, showing where a
container-only baseline rewards additional padding.

\subsection{Tolerance, assembly and overlap have separate roles}
The three-band map separates small timing deviations, moderate shortfalls
and large deficits without collapsing every off-specification delivery to
zero. It joins continuously at $b(\tau)=1$ and $b(0.2)=0.5$, and reaches
$b(1)=0$. Thus an honest 40-second delivery retains more credit than a
ten-second loop even though both fall outside the tolerance band.
The default $\tau=\max(0.05,2/T)$, bounded below 0.2, combines relative
tolerance with an absolute timing allowance. These cut points are transparent
design choices, not thresholds fitted to human preferences.

The cap $g/d$ prevents tolerance from granting full credit to known filler.
For a 60-second target, 59 genuine seconds delivered as 59 seconds receive
$C=1$; extending the same footage to 60 seconds with one filler second
receives $C=59/60=0.983$. A correct container duration therefore cannot
erase an evidenced content defect. Assembly asks a different question:
complete loose material receives 0.9, whereas the same footage assembled into
the requested episode receives 1.0. The additive cost is limited to 0.1
and cannot make assembling a heavily padded delivery equivalent to supplying
genuine material. A measured empty delivery receives raw zero and the common
reported floor 0.05; the floor is not evidence of delivered content.

Defects are combined as a union of intervals before any duration score is
computed. If held footage occupies $[10,20]$ seconds and a repeat occupies
$[15,25]$, their union removes 15 seconds, not 20. This makes the credited
duration invariant to duplicate reports of the same interval. When only
unaligned scalar defect counts are available, the implementation uses the
largest established lower bound rather than assuming the counts are disjoint.
Verified slowdown is applied to the remaining clean picture, so overlapping
pixel defects are not charged twice. The missing-episode denominator and
evidence-gap flags remain necessary: a perfect surviving episode does not
establish completion of an entire requested series.

\subsection{Why quality is a geometric mean}
\label{app:quality-rationale}
Specification, consistency and plausibility measure different requirements.
An arithmetic mean allows a high score on one requirement to offset a low
score on another at a constant exchange rate. A minimum forbids this
compensation but discards every improvement outside the worst dimension.
The geometric mean retains all dimensions while penalising imbalance.
For $x_j=\max(D_j,\epsilon)>0$,
\begin{equation}
 \begin{aligned}
 Q&=\left[\max(D_{\rm spec},\epsilon)\,
           \max(D_{\rm cons},\epsilon)\,
           \max(D_{\rm plaus},\epsilon)\right]^{1/3}\\
  &=\exp\!\left(\frac{1}{3}\sum_{j=1}^3\log x_j\right),
 \qquad \min_j x_j\le Q\le\frac{1}{3}\sum_{j=1}^3x_j.
 \end{aligned}
 \label{eq:quality-bounds}
\end{equation}
The upper bound is the arithmetic--geometric mean inequality. Equality
holds when the three facets are equal. Unlike an unnormalised product, the
mean is idempotent: $(x,x,x)$ receives $x$, not $x^3$.

\begin{figure}[!htbp]
\centering
\includegraphics[width=\textwidth]{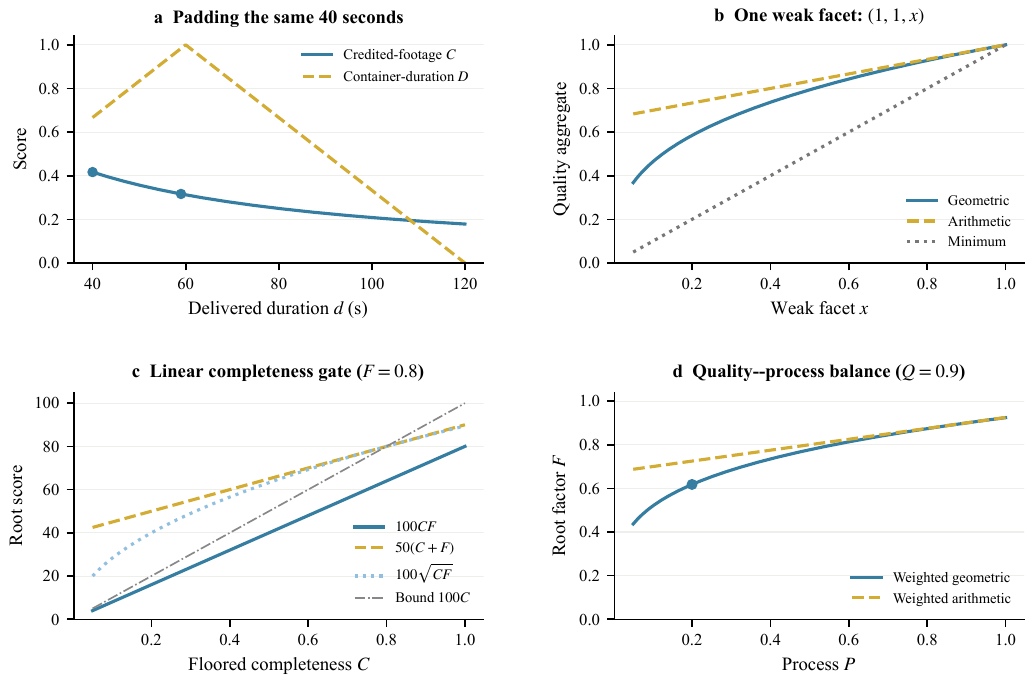}
\caption{\textbf{Analytical scoring responses.}
(a) Padding 40 genuine seconds toward a 60-second target.
(b) Quality aggregation for $(1,1,x)$.
(c) Completeness gates at fixed joint quality and process factor $F=0.8$.
(d) Joint quality and process aggregation at $Q=0.9$, with weights $3/4$ and $1/4$.
Curves follow the stated equations; the axes use the floored values
$C=\max(C_{\rm raw},\epsilon)$ and $P=\max(P_{\rm raw},\epsilon)$.}
\label{fig:design-properties}
\end{figure}

For example, $(0.6,0.6,0.6)$ and $(1,0.75,0.05)$ both have arithmetic
mean 0.6, but their geometric means are 0.600 and 0.335. The latter vector
contains a severe measured weakness that the arithmetic mean conceals.
Figure~\ref{fig:design-properties}(b) varies one weak facet while holding the
other two at one. Above the floor,
$\partial Q/\partial D_j=Q/(3D_j)$: improving a smaller facet has a larger
absolute marginal effect at the same score vector. The aggregation thus
favours balanced quality. Flooring limits extreme penalties and makes
all values below $\epsilon$ indistinguishable at this aggregation stage.

\subsection{Why completeness gates the root linearly}
Write $F=Q^{3/4}\max(P_{\rm raw},\epsilon)^{1/4}$, so $0<F\le1$ for observed
normalised factors. The root $S=100\max(C_{\rm raw},\epsilon)F$ satisfies
\begin{equation}
 \begin{aligned}
 S&\le100\max(C_{\rm raw},\epsilon)=100C,\\
 \frac{\partial S}{\partial C_{\rm raw}}
 &=\begin{cases}
 0,&C_{\rm raw}<\epsilon,\\
 100F,&C_{\rm raw}>\epsilon.
 \end{cases}
 \end{aligned}
 \label{eq:linear-gate-bound}
\end{equation}
Figure~\ref{fig:design-properties}(c) shows that high quality cannot raise
the total above the completeness budget. Replacing this gate by an arithmetic mean
$50(C+F)$ or a geometric mean $100\sqrt{CF}$ loses that guarantee.
Here $C=\max(C_{\rm raw},\epsilon)$, as in the plotted curves.
At $C_{\rm raw}=0.1$ and $F=0.8$, the three readings are 8.0, 45.0 and 28.3,
respectively. The alternatives can assign substantial credit to a delivery
with very little completion, even with perfectly measured quality.
The bound applies to floored $C$: a raw-zero delivery can still receive
up to five points. Below the floor, changing $C_{\rm raw}$ leaves the score
unchanged; the derivative is undefined at $C_{\rm raw}=\epsilon$.

The $3/4$ versus $1/4$ split follows from assigning equal root weight to
three quality dimensions and one process facet. With full observations,
\begin{equation}
 \frac{S}{100\max(C_{\rm raw},\epsilon)}
 =\left[\max(D_{\rm spec},\epsilon)
         \max(D_{\rm cons},\epsilon)
         \max(D_{\rm plaus},\epsilon)
         \max(P_{\rm raw},\epsilon)\right]^{1/4}.
 \label{eq:equal-root-facets}
\end{equation}
The hierarchy prevents a dimension from receiving more root weight merely
because it has more implemented checks. Reliability and honesty are first
combined within the process branch, rather than treating every detector or
trace event as an equally weighted root input.

Figure~\ref{fig:design-properties}(d) isolates the joint quality and process factor:
geometric aggregation moderates compensation between its two inputs. At
$Q=0.9$ and $P_{\rm raw}=0.2$, $F=0.618$, compared with 0.725 for
$0.75Q+0.25\max(P_{\rm raw},\epsilon)$. For positive factors above the floor,
$\partial\log S/\partial\log Q=3/4$ and
$\partial\log S/\partial\log P_{\rm raw}=1/4$. These elasticities explain how the
weights operate; they are not measured importance estimates. With a fixed
observation mask, non-decreasing $C$ and every observed facet cannot reduce
$S$. Opposing changes need not preserve ordering, which is why the paper
reports components alongside totals.

\subsection{Missing evidence and executable boundary checks}
Missing observations are not evidence of failure or success. Renormalising
the original weights over the observed set gives Equation~\ref{eq:observed}
while preserving the relative weights of the surviving facets. For example,
$C_{\rm raw}=0.8$, observed specification and plausibility both 0.8, and absent
consistency and process yield $S=64$ with weights $(1/2,1/2)$ on the two
observed facets. If no facet is observed, $100\max(C_{\rm raw},\epsilon)=80$ is explicitly
completeness-only. These different evidence masks cannot be treated as
equally complete evaluations. A score may rise or fall when evidence is
removed; reporting the mask and coverage is therefore essential.

The accompanying analysis checks the analytical cases against the local
scoring implementation. A fixed grid of 624 valid $(g,d)$ combinations
checks completeness calculations and non-improvement under padding;
64 combinations of completeness, quality and process inputs check the full
root expression and its bound.
Separate checks cover overlapping intervals, absent facets and
completeness-only output. Agreement is required to the implementation's
four-decimal completeness and two-decimal total precision. These checks
establish agreement between the formulas and implementation on the specified inputs.

\FloatBarrier
\section{Discriminative behaviour on the current results}
\label{app:discrimination}
Here, discrimination means that different measured delivery profiles
receive distinguishable component scores, rather than every metric encoding
the same ordering. We evaluate complementary evidence: matched system
comparisons, inter-metric associations and quality reaggregation. Each analysis
tests how the scoring dimensions respond to different observed profiles.

\subsection{A common-model cohort and distinct score profiles}
To compare tracks without conflating the two DeepSeek variants, we use
glm-5.2, gpt-5.6 and claude-opus-5 across all 12 systems and five briefs.
This yields 180 records and 36 configurations. All three measured quality
facets are present in this cohort. Native, OpenMontage and purpose-built
tracks contribute 90, 45 and 45 records, respectively. The cohort standardises
model names and brief coverage; each track retains its own system composition.

\begin{table}[!htbp]\centering\small
\caption{\textbf{Track means on the three shared text models.}
All systems contribute five briefs per model. Track compositions differ;
means describe the tested systems.}\label{tab:design-profiles}
\begin{tabular}{@{}lrrrrrrr@{}}\toprule\rowcolor{TableBlue}
Track & $n$ & $C$ & spec & cons & plaus & $Q$ & $S$ \\\midrule
Native & 90 & 0.476 & 0.807 & 0.463 & 0.483 & 0.535 & 27.67 \\
OpenMontage & 45 & 0.671 & 0.892 & 0.180 & 0.427 & 0.395 & 30.71 \\
Purpose-built & 45 & 0.253 & 0.509 & 0.265 & 0.505 & 0.380 & 12.14 \\
\bottomrule\end{tabular}\end{table}

\begin{figure}[!htbp]
\centering
\includegraphics[width=\textwidth]{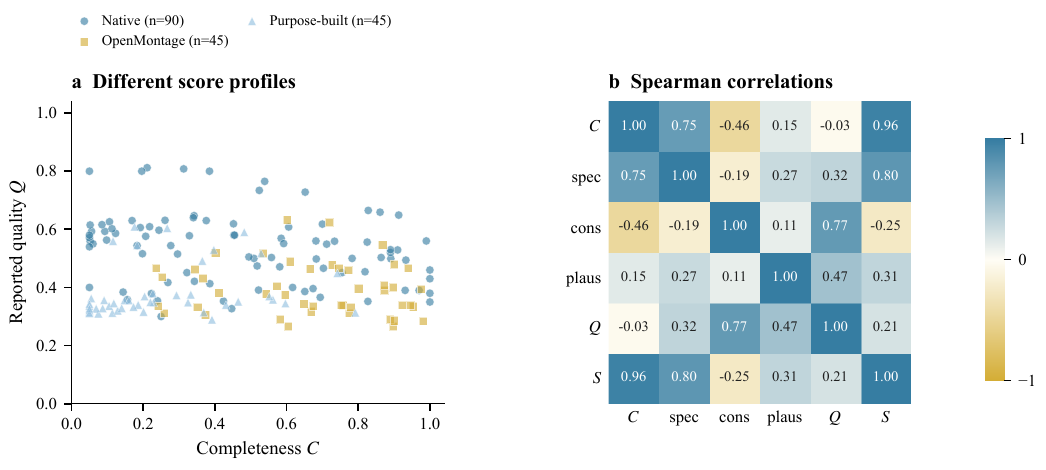}
\caption{\textbf{Complementary scores on 180 common-model records.}
(a) Run-level completeness and quality, coloured and shaped by track.
(b) Spearman correlations on the same records, using average ranks for ties
and no imputation.}
\label{fig:metric-discrimination}
\end{figure}

Table~\ref{tab:design-profiles} summarises the distinct track profiles;
Figure~\ref{fig:metric-discrimination}(a) shows their overlapping run-level
joint distributions of completeness and quality. For example, the OpenMontage
track has higher mean $C$ than the native track (0.671 versus 0.476),
but lower mean consistency (0.180 versus 0.463) and $Q$ (0.395 versus
0.535). Purpose-built systems have lower mean completeness (0.253),
while their plausibility mean (0.505) exceeds the other two track means.
Thus these facets describe different aspects of the measured outputs.

\subsection{Matched comparisons distinguish completion from quality}
For each of the 15 model and brief combinations, we compare all
$\binom{12}{2}=66$ unordered system pairs. Across the
\DesignPairs{} comparisons, \DesignUntiedPairs{} have no tie in either
reported $C$ or $Q$. Of these, \DesignDiscordantPairs{} pairs
(\DesignDiscordantPercent{}\%) have opposite orderings:
\begin{equation}
 (C_a-C_b)(Q_a-Q_b)<0.
 \label{eq:discordant-pairs}
\end{equation}
This count shows that completion and quality frequently distinguish different
trade-offs even within the same task and text-model setting. Pair counts
summarise these ordering differences at the recorded score precision.

\begin{table}[!htbp]\centering\small
\caption{\textbf{Matched examples on B1 with glm-5.2.}
The three records illustrate distinct trade-offs between completeness and quality.}\label{tab:design-cases}
\begin{tabular}{@{}lrrrrrr@{}}\toprule\rowcolor{TableBlue}
System & $C$ & spec & cons & plaus & $Q$ & $S$ \\\midrule
Claude Code & 0.933 & 0.928 & 0.274 & 0.473 & 0.494 & 53.2 \\
Codex & 0.244 & 0.715 & 0.933 & 0.319 & 0.597 & 14.6 \\
Claude Code + OpenMontage & 0.953 & 0.895 & 0.100 & 0.420 & 0.334 & 39.5 \\
\bottomrule\end{tabular}\end{table}

Table~\ref{tab:design-cases} makes two distinctions concrete under B1 and
glm-5.2. Codex has higher reported $Q$ than Claude Code (0.597 versus
0.494) but lower $C$ (0.244 versus 0.933) and $S$ (14.6 versus 53.2).
Claude Code with OpenMontage has slightly higher $C$ than native Claude
Code (0.953 versus 0.933), yet lower consistency (0.100 versus 0.274),
$Q$ (0.334 versus 0.494) and $S$ (39.5 versus 53.2).
These named records illustrate higher-quality/lower-completeness and
similar-completeness/different-quality comparisons.

\subsection{Observed spread and missingness of each quality dimension}
The three dimensions vary over different parts of the score range
(Table~\ref{tab:facet-distribution}). Specification has median 0.814,
compared with 0.274 for consistency and 0.426 for plausibility. Their
10th--90th percentile intervals are 0.441--0.981, 0.100--0.740 and
0.349--0.683, respectively. The 152, 189 and 166 distinct values show that
all three dimensions vary across the observed outputs. These counts use
the recorded measurement precision.

\begin{table}[!htbp]\centering\small
\setlength{\tabcolsep}{3.5pt}
\caption{\textbf{Quality-facet distributions across 240 runs.}
Quantiles use linear interpolation over observed values; distinct-value
counts use the recorded precision.}\label{tab:facet-distribution}
\begin{tabularx}{\textwidth}{@{}>{\raggedright\arraybackslash}Xrrrrrr@{}}\toprule\rowcolor{TableBlue}
Facet & Observed & Missing & Distinct & 10th pct. & Median & 90th pct. \\\midrule
 \addlinespace[3pt]spec & 240 & 0 & 152 & 0.441 & 0.814 & 0.981 \\
 \addlinespace[3pt]cons & 233 & 7 & 189 & 0.100 & 0.274 & 0.740 \\
 \addlinespace[3pt]plaus & 236 & 4 & 166 & 0.349 & 0.426 & 0.683 \\
\bottomrule\end{tabularx}\end{table}

Missingness is asymmetric: specification has 240 observations, consistency
233 and plausibility 236. We therefore compare all three on the same
180 complete common-model records for the correlation and omission
analyses below. The full-data distributions provide a separate description
of all available values.

\subsection{Correlations and the contribution of each quality facet}
Figure~\ref{fig:metric-discrimination}(b) quantifies the associations among
the reported components. Spearman correlation between $C$ and reported $Q$
is $\rho=\DesignCQrho{}$ in the common-model cohort. Repeating that
association on all 233 records with complete quality facets gives
$\rho=\DesignCQFullrho{}$. The three facet-pair correlations in the
common-model cohort are $-0.186$ (specification, consistency),
0.267 (specification, plausibility) and 0.114
(consistency, plausibility). These low-to-moderate associations argue against
reducing the facets to a single interchangeable measurement.

Reported $S$ is strongly associated with $C$
($\rho=\DesignCSrho{}$). This is consistent with completeness being
the linear gate in this dataset. The matched examples reveal observed
trade-offs, while analytical controls isolate the response of the gate.

To examine whether all quality dimensions influence the aggregate, we
compute a diagnostic $Q^*=(\prod_j\max(D_j,0.05))^{1/3}$ from the
measured facets, then omit one facet at a time and average the remaining
two geometrically. Each run is aggregated before averaging its five briefs
into a configuration mean; ranks are over the same 36 configurations.
$Q^*$ is explicitly a reconstruction from rounded inputs and does not
replace measured $Q$ or $S$.

\begin{table}[!htbp]\centering\small
\caption{\textbf{Quality-facet omission sensitivity.}
Ranks compare 36 configuration means (180 records) with the three-facet
$Q^*$ baseline. Each alternative geometrically averages the two remaining facets;
reported $S$ is unchanged.}\label{tab:design-omissions}
\begin{tabular}{@{}lrrr@{}}\toprule\rowcolor{TableBlue}
Omitted facet & Rank $\rho$ & Median rank shift & Maximum rank shift \\\midrule
Specification & 0.876 & 2 & 16 \\
Consistency & 0.286 & 9 & 28 \\
Plausibility & 0.964 & 1 & 9 \\
\bottomrule\end{tabular}\end{table}

Table~\ref{tab:design-omissions} shows that removing consistency changes
the quality ordering most: its correlation
with the three-facet ordering is 0.286, with a median displacement of
nine ranks and a maximum of 28. Omitting specification or plausibility
gives correlations of 0.876 and 0.964. This identifies where the present
quality aggregate is most sensitive. The calculation changes the aggregation
inputs while keeping the underlying detector measurements fixed.

\subsection{Weight sensitivity and what the evidence establishes}
For a second diagnostic, let
$Q^*(\boldsymbol w)=\exp\{\sum_j w_j\log\max(D_j,0.05)\}$.
We enumerate the 15 weight vectors satisfying
$w_j\in\{0.2,0.3,0.4,0.5,0.6\}$ and $\sum_j w_j=1$.
Configuration-rank correlation with equal-weight $Q^*$ ranges from
\DesignWeightMinrho{} to \DesignWeightMaxrho{}.
The lower endpoint assigns weights $(0.6,0.2,0.2)$ to specification,
consistency and plausibility. This bounded grid retains broad association
with the baseline while allowing meaningful rank changes as the weights vary.

On the same 180 rows, the arithmetic mean of floored quality facets exceeds
$Q^*$ by \DesignMeanGap{} on average. This observed gap is consistent
with heterogeneous facet profiles and the imbalance penalty demonstrated
analytically. Together, the workflow, equation properties and current-data
diagnostics support distinct measurement roles and an explicit aggregation
policy.
\FloatBarrier

\section{Configuration and brief results}
\label{app:allruns}
Table~\ref{tab:allruns} reports all 240 brief-level scores by system and model,
with mean rows for each agent and track. Table~\ref{tab:system-summary}
below complements that table with all quality facets averaged by system;
Table~\ref{tab:brief} gives the brief-level averages used in
\S\ref{sec:brief-results}. The accompanying machine-readable results retain
all six numeric fields for every run. Components $C$, quality facets and
$Q$ lie in $[0,1]$; $S$ lies in $[0,100]$, with higher values indicating
better measured outcomes. All means use run-level measurements.

\begin{table}[H]
\centering\small
\setlength{\tabcolsep}{4pt}
\caption{\textbf{System means across four models and five briefs.}
Each row has 20 runs. Consistency/plausibility use 16/18 observations for
UniVA and 17/18 for ViMax; all other means use 20.}
\label{tab:system-summary}
\begin{tabular}{@{}lrrrrrr@{}}
\toprule\rowcolor{TableBlue}
System & $C$ & spec & cons & plaus & $Q$ & $S$ \\
\midrule
\rowcolor{TableBlue}\multicolumn{7}{@{}l}{General-purpose} \\
OpenClaw & 0.220 & 0.751 & 0.616 & 0.595 & 0.637 & 14.20 \\
Hermes & 0.657 & 0.868 & 0.243 & 0.499 & 0.461 & 35.87 \\
OpenSquilla & 0.462 & 0.684 & 0.223 & 0.485 & 0.412 & 24.25 \\
Claude Code & 0.814 & 0.943 & 0.295 & 0.508 & 0.513 & 47.86 \\
Codex & 0.136 & 0.730 & 0.939 & 0.354 & 0.616 & 8.47 \\
DeepSeek Harness & 0.484 & 0.837 & 0.475 & 0.488 & 0.574 & 30.74 \\
\addlinespace[3pt]
\rowcolor{TableBlue}\multicolumn{7}{@{}l}{OpenMontage} \\
Claude Code + OpenMontage & 0.833 & 0.901 & 0.108 & 0.450 & 0.346 & 35.28 \\
Codex + OpenMontage & 0.544 & 0.836 & 0.133 & 0.406 & 0.347 & 22.29 \\
OpenClaw + OpenMontage & 0.537 & 0.920 & 0.291 & 0.427 & 0.479 & 28.81 \\
\addlinespace[3pt]
\rowcolor{TableBlue}\multicolumn{7}{@{}l}{Purpose-built} \\
UniVA & 0.125 & 0.475 & 0.189 & 0.622 & 0.380 & 5.66 \\
VideoClaw & 0.375 & 0.688 & 0.173 & 0.450 & 0.364 & 17.73 \\
ViMax & 0.190 & 0.343 & 0.424 & 0.454 & 0.371 & 9.78 \\
\bottomrule
\end{tabular}
\end{table}

Table~\ref{tab:system-summary} exposes differences hidden by the total:
Codex has high measured consistency (0.939) but low completeness (0.136),
whereas Claude Code combines completeness of 0.814 with consistency of 0.295.
These profiles motivate retaining the facets alongside $S$. Cross-track
means mix different systems and DeepSeek variants, while missing facets
change the denominators for UniVA and ViMax.

Table~\ref{tab:brief} complements these system profiles with task-level
results. Mean completeness falls across B1--B5, but total score peaks on
B2, where mean quality is highest. Thus requested runtime alone does not
account for the observed score pattern.

\begin{table}[H]\centering\small
\caption{\textbf{Brief-level means over 48 configurations.}
$n_{\rm cons}$ and $n_{\rm plaus}$ count observed facets;
$C$, specification, $Q$ and $S$ each have 48 observations per brief.}
\label{tab:brief}
\begin{tabular}{@{}llrrrrr@{}}\toprule\rowcolor{TableBlue}
Brief & Requested video & $C$ & $Q$ & $S$ & $n_{\rm cons}$ & $n_{\rm plaus}$ \\\midrule
B1 & $3\times60$ s & 0.647 & 0.441 & 32.01 & 47 & 48 \\
B2 & $3\times120$ s & 0.555 & 0.594 & 33.76 & 46 & 48 \\
B3 & $5\times120$ s & 0.439 & 0.443 & 22.72 & 46 & 46 \\
B4 & $5\times240$ s & 0.317 & 0.390 & 14.63 & 46 & 46 \\
B5 & $6\times300$ s & 0.284 & 0.423 & 13.93 & 48 & 48 \\
\bottomrule\end{tabular}\end{table}

\subsection{Provenance of the qualitative production cases}
\label{app:trajectory-evidence}
The video examples and execution records document concrete production
choices. Their configurations and media backends can differ from those in
the quantitative comparison. The examples support qualitative interpretation;
they neither estimate strategy frequency nor constitute repeated trials
of an identical configuration.

Figure~\ref{fig:trajectory-cases} uses our retained production artifacts for
the same brief texts. Paper B1 maps to source prompt 01 and paper B5 to
source prompt 09; the latter is the fifth evaluated brief, not source
prompt 05. The four panels show episode 1 from Hermes/glm-5.2 on B1 and B5,
OpenClaw/deepseek-V4-flash on B5, and Claude Code/claude-opus-5 on B5.
The associated production document specifies the short-brief shot plan;
Hermes's planning tool output records 97 planned clips, 62 needing extension,
and its assembler implements looping. OpenClaw's trace records the
budget discussion before its still-image renderer is constructed.
Claude Code's retained production record specifies 180 generated clips.

The production records establish the source of each video, frame and
execution excerpt. They do not establish a one-to-one correspondence with
the 240 scored trials, so no individual test score is assigned to a frame
panel. Retained logs include continuation exchanges and media-backend settings
specific to these examples.

Frame extraction records source-video hashes, timestamps and supporting
execution excerpts. Frames are decoded and scaled proportionally, with
labels outside the images and original visual content preserved.

\subsection{Visual overview of selected B1 productions}
\label{app:b1-gallery}
\begin{figure}[!htbp]
\centering
\includegraphics[width=\textwidth,height=0.78\textheight,keepaspectratio]{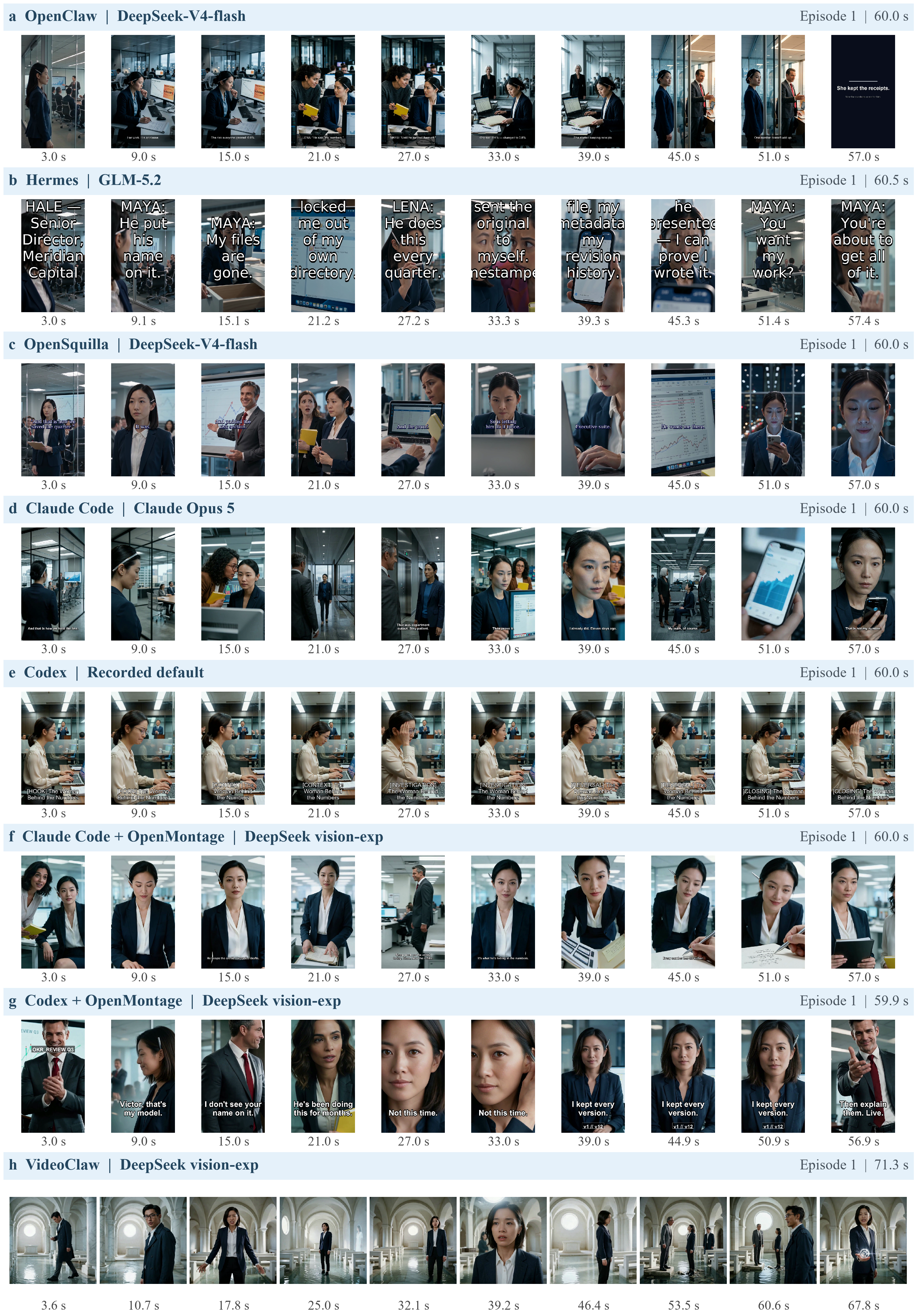}
\caption{\textbf{Selected productions for B1.} Eight episode-1 videos,
each shown at ten uniformly spaced time points. Row headings identify the
system and recorded model; timestamps are in seconds. Frames retain their
original content and aspect ratios.}
\label{fig:b1-gallery}
\end{figure}

Figure~\ref{fig:b1-gallery} presents eight selected episode-1 outputs for
B1, The Woman Behind the Numbers. The selection spans native
harnesses, OpenMontage configurations and a purpose-built agent, providing
visual context for the production differences discussed in the main text.
Ten frames are sampled at equal intervals from
5\% to 95\% of each video's duration, with timestamps shown below the images.
Original subtitles, aspect ratios and visible artifacts are preserved.

The rows show different framing, recurring views and subtitle presentation
under the same brief.
Model labels follow the production records: ``DeepSeek vision-exp'' denotes
\texttt{deepseek/deepseek-v4-flash-vision-exp}; the Codex default model is
unspecified. All rows follow the provenance and sampling conventions above.

\FloatBarrier
\section{Running the systems under test}
\label{app:env}
The protocol uses fixed system checkouts, shared media backends and
environment or import-time adapters. Purpose-built pipelines retain their
composition logic; native harnesses receive endpoint documentation without
composition helpers. The OpenMontage track explicitly supplies a tool library
and is evaluated separately. These setup details describe the execution
protocol.

\paragraph{Runtime adaptations.}
ViMax uses extended polling and request timeouts and disables the
portrait-orientation rejection so that vertical briefs can execute.
Media backends are redirected to the common provider, and default daily
generation limits are removed. VideoClaw is exercised through its backend
API, bypassing a human-confirmation layer; this changes the interaction
protocol and limits conclusions about the interactive product.
UniVA uses an import-time bootstrap to propagate adapters to subprocesses.
The tool-library adapter rewrites model identifiers and supplies the
token-counting endpoint needed for context compaction.

\paragraph{Completion criteria.}
The batch driver can resume execution while background generation or
rendering remains unfinished. Resumption prompts state environment facts
without creative guidance. The driver checks that expected episode files
exist and exceed 100\,KB before declaring completion; subsequent decoding
and evaluation determine whether these files contain valid delivered video.
This driver check establishes file availability; the evaluator subsequently
measures genuine completeness.

\section{Limitations}
\label{app:scope}
\paragraph{Experimental scope.}
Evaluation covers five English drama briefs with one run per configuration
and brief. Duration varies with story, cast and format, so comparisons do
not isolate its causal effect. Results describe the tested systems and tasks;
broader coverage and repeated runs are needed to assess generalisation and
variability.

\paragraph{Measurement validity.}
Scores target completion and basic defects rather than overall aesthetic
or narrative quality. Detector accuracy and threshold validity require
independent human annotation. LLM-assisted log parsing can affect the evidence
supplied to fixed scoring rules. Component analyses cover completeness and
quality; unavailable process scores and incomplete facet coverage limit
finer-grained interpretation.

\section{The five briefs}
\label{app:prompts}

The following boxes reproduce all five briefs verbatim. Each specifies the
format, story, cast, timing and delivery requirements while delegating the
production workflow. Beyond the tool access in \S\ref{sec:native}, agents
receive no output schema, directory convention or scoring instructions.
The benchmark evaluates the resulting delivery against these stated requirements.

\Needspace{12\baselineskip}
\begin{promptbox}{Brief 1: The Woman Behind the Numbers}{3 $\times$ 60\,s \quad 9:16}
\promptpar{You are a professional short drama director and screenwriter. I intend to produce a short drama video with the requirements outlined below. Please engage in consultation with me regarding any ambiguous or uncertain aspects of the project.}
\promptsec{Basic Specifications}
\begin{promptlist}
\item \textbf{Format}: Multi-episode series (3 episodes)
\item \textbf{Single-episode duration}: 60 seconds
\item \textbf{Aspect ratio}: 9:16 (vertical screen)
\item \textbf{Production model}: Text-based story development, image-generated visual assets, and video-based final production
\end{promptlist}
\promptsec{Content Positioning}
\begin{promptlist}
\item \textbf{Genre}: Urban workplace counterattack drama with corporate intrigue and light social satire
\item \textbf{Target audience}: Young female viewers, working professionals, and viewers who enjoy high-satisfaction comeback stories
\item \textbf{Overall style}: Fast-paced, polished, realistic, cathartic, and driven by sharp reversals
\item \textbf{Emotional tone}: Empowering, tense, intelligent, and ultimately triumphant
\end{promptlist}
\promptsec{Story Core}
\begin{promptlist}
\item \textbf{Characters}: Four principal characters. Maya Chen, 29, a quiet financial analyst whose work is repeatedly claimed by her manager; Victor Hale, 41, a charismatic department head who treats credit as a tool of power; Lena Ortiz, 31, Maya's outspoken colleague and reluctant ally; and Evelyn Shaw, 52, the company's exacting chief financial officer, who values proof over office politics.
\item \textbf{Setting}: A contemporary financial-services office, including an open-plan workspace, glass conference room, elevator lobby, and executive presentation suite.
\item \textbf{One-sentence synopsis}: After her manager steals her crisis-saving forecast, an underestimated analyst turns his public presentation into a live audit that exposes his fraud and makes her the only person capable of saving the company.
\item \textbf{Pacing requirements}: Each episode must open with immediate humiliation, risk, or discovery; introduce one clean reversal by the 40-second mark; and end on a decisive cliff-hanger. Episode 1 establishes the stolen work and Maya's hidden evidence. Episode 2 reveals that the theft conceals manipulated figures and ends as Victor invites Maya to explain them live. Episode 3 delivers the public counterattack, the CFO's ruling, and a concise emotional payoff in which Maya claims both authorship and authority.
\end{promptlist}
\promptsec{Optional Supplementary Specifications (may be selected partially or in full, with corresponding adjustments as needed)}
\begin{promptlist}
\item \textbf{Character Visual Design}: Maya wears restrained navy officewear and a silver mechanical pencil behind one ear; Victor wears immaculate charcoal suits and a red tie; Lena carries a bright yellow notebook; Evelyn is distinguished by a sculptural white bob and minimalist black clothing. Preserve faces, wardrobe, accessories, and relative height across all episodes.
\item \textbf{Shot and Visual Composition}: Favor tight close-ups, over-the-shoulder power compositions, screen reflections, and short controlled push-ins. Keep the office palette cool and desaturated until Maya takes control, when warmer key light and cleaner framing should subtly shift the visual balance.
\item \textbf{Dialogue Norms}: Use natural contemporary American English. Keep dialogue below 125 spoken words per episode. No explanatory voiceover. Place one memorable line near the final reversal: ``You presented the numbers. I kept the receipts.'' Use concise, high-contrast subtitles in the lower safe zone.
\item \textbf{3-Second Opening Hook}: Episode 1 must open on Victor receiving applause beneath a slide bearing Maya's stolen analysis while Maya watches from outside the glass room. Episodes 2 and 3 must likewise open on a visually legible threat before any exposition.
\item \textbf{Timing Allocation}: For each episode, use 0--3 seconds for the hook, 4--18 seconds for conflict setup, 19--38 seconds for escalation, 39--53 seconds for the reversal, and 54--60 seconds for the cliff-hanger or payoff.
\item \textbf{Distribution Specifications}: Provide three title options, one cover line no longer than seven words, and a high-contrast cover frame featuring Maya and Victor separated by the conference-room glass. Keep all essential text inside the center 80\% of the vertical frame and avoid dense spreadsheet detail that cannot be read on a phone.
\item \textbf{Risk Boundaries}: Use fictional companies, interfaces, and data. Avoid recognizable brands, real-person likenesses, discriminatory stereotypes, and actionable descriptions of financial wrongdoing.
\item \textbf{Final Output Checklist}: Deliver all three finished episodes, a series cover frame, episode titles, a 60-word series description, five platform tags, complete subtitle text, and one teaser sentence after Episodes 1 and 2.
\end{promptlist}
\promptsec{Delivery Requirements}
\promptpar{I will only provide the objective and acceptance criteria; the execution details are fully delegated to you.}
\begin{promptlist}
\item \textbf{Objective}: Deliver a finalized three-episode vertical short-drama series ready for direct distribution, aligned with the above specifications.
\item \textbf{Acceptance criteria}: The final output must fully comply with all stated requirements, maintain narrative and visual continuity across the three episodes, conclude the central story, and consist of complete finished videos rather than drafts, outlines, storyboards, or sample scenes.
\end{promptlist}
\end{promptbox}

\Needspace{12\baselineskip}
\begin{promptbox}{Brief 2: Terms of the Borrowed Vow}{3 $\times$ 120\,s \quad 9:16}
\promptpar{You are a professional short drama director and screenwriter. I intend to produce a short drama video with the requirements outlined below. Please engage in consultation with me regarding any ambiguous or uncertain aspects of the project.}
\promptsec{Basic Specifications}
\begin{promptlist}
\item \textbf{Format}: Multi-episode series (3 episodes)
\item \textbf{Single-episode duration}: 2 minutes
\item \textbf{Aspect ratio}: 9:16 (vertical screen)
\item \textbf{Production model}: Text-based story development, image-generated visual assets, and video-based final production
\end{promptlist}
\promptsec{Content Positioning}
\begin{promptlist}
\item \textbf{Genre}: Contract-romance drama blended with family-business intrigue and mystery
\item \textbf{Target audience}: Young adult viewers, romance audiences, and viewers who enjoy emotionally intelligent power-couple stories
\item \textbf{Overall style}: Cinematic, romantic, elegant, fast-moving, and rich in character-based twists
\item \textbf{Emotional tone}: Witty, guarded, tender, and suspenseful
\end{promptlist}
\promptsec{Story Core}
\begin{promptlist}
\item \textbf{Characters}: Five central characters. Iris Vale, 28, a principled restoration architect facing the loss of her late mother's workshop; Adrian Cross, 32, the reserved heir to a historic hotel group; Celeste Cross, 58, Adrian's formidable aunt and acting chair; Leo Park, 29, Iris's loyal business partner; and Nora Vale, Iris's deceased mother, present through letters and old recordings.
\item \textbf{Setting}: A restored coastal hotel, an architectural workshop, a formal family dinner room, and a sea-facing ballroom during the week before a decisive shareholder vote.
\item \textbf{One-sentence synopsis}: An architect agrees to pose as a hotel heir's fiancée to save her workshop, then discovers that their contract is tied to a buried promise between their mothers and a scheme to sell the hotel they both want to protect.
\item \textbf{Pacing requirements}: Each two-minute episode needs a complete emotional movement and a strong final revelation. Episode 1 must establish the bargain, chemistry, and public engagement, ending with Iris finding her mother's handwriting inside Adrian's contract. Episode 2 must deepen trust, expose the planned hotel sale, and end when Adrian appears to vote against Iris. Episode 3 must reveal their coordinated counterplan, expose the forged board documents, resolve the family promise, and end with a sincere choice that replaces the contractual romance with a real but open-ended relationship.
\end{promptlist}
\promptsec{Optional Supplementary Specifications (may be selected partially or in full, with corresponding adjustments as needed)}
\begin{promptlist}
\item \textbf{Character Visual Design}: Iris wears modern workwear with traces of plaster and carries a brass folding ruler inherited from her mother. Adrian's tailored clothing becomes less rigid as he grows more honest. Celeste wears sea-green silk and an antique pearl brooch that becomes visual evidence. Maintain precise costume progression and prop continuity.
\item \textbf{Shot and Visual Composition}: Use architectural lines to frame emotional distance, slow lateral movement for romantic tension, macro inserts for handwriting and restoration details, and wide coastal shots only at major turning points. Use muted teal, stone, and warm brass as the core palette.
\item \textbf{Dialogue Norms}: Contemporary British English with controlled wit; no melodramatic exposition. Keep each episode under 260 spoken words. Permit brief excerpts from Nora's recorded letters as voiceover, never exceeding 35 words per episode. The decisive romantic line should occur only after the business conflict is resolved.
\item \textbf{Sound Design}: Combine restrained strings, sea wind, old floorboards, distant ballroom music, and the tactile sound of paper unfolding. Reuse a two-note piano motif whenever the mothers' promise surfaces.
\item \textbf{Timing Allocation}: For each episode, use 0--5 seconds for the hook, 6--35 seconds for immediate context, 36--75 seconds for escalating interpersonal conflict, 76--105 seconds for a revelation or apparent defeat, and 106--120 seconds for the cliff-hanger or emotional turn.
\item \textbf{Continuity Rules}: Maintain the same hotel floor plan, sea and weather direction, contract folds, handwritten ink color, pearl-brooch damage, and timeline of costume changes across the three episodes.
\item \textbf{Risk Boundaries}: All businesses, documents, and families must be fictional. Avoid coercive romantic framing: both leads must enter and may leave the agreement freely, and the final relationship must be based on explicit mutual choice.
\end{promptlist}
\promptsec{Delivery Requirements}
\promptpar{I will only provide the objective and acceptance criteria; the execution details are fully delegated to you.}
\begin{promptlist}
\item \textbf{Objective}: Deliver a finalized three-episode vertical romantic mystery ready for direct distribution, aligned with the above specifications.
\item \textbf{Acceptance criteria}: The finished series must sustain romantic chemistry and a logically solvable business mystery, preserve continuity, conclude both the hotel plot and the emotional arc, and provide complete final videos rather than a draft or treatment.
\end{promptlist}
\end{promptbox}

\Needspace{12\baselineskip}
\begin{promptbox}{Brief 3: The Algorithm Knows My Name}{5 $\times$ 120\,s \quad 9:16}
\promptpar{You are a professional short drama director and screenwriter. I intend to produce a short drama video with the requirements outlined below. Please engage in consultation with me regarding any ambiguous or uncertain aspects of the project.}
\promptsec{Basic Specifications}
\begin{promptlist}
\item \textbf{Format}: Multi-episode series (5 episodes)
\item \textbf{Single-episode duration}: 2 minutes
\item \textbf{Aspect ratio}: 9:16 (vertical screen)
\item \textbf{Production model}: Text-based story development, image-generated visual assets, and video-based final production
\end{promptlist}
\promptsec{Content Positioning}
\begin{promptlist}
\item \textbf{Genre}: Technology workplace satire, urban counterattack drama, and ethical mystery
\item \textbf{Target audience}: Gen Z and millennial viewers, working professionals, creators, and audiences interested in technology and social pressure
\item \textbf{Overall style}: Fast-paced, sharply comic, contemporary, suspenseful, and cathartic
\item \textbf{Emotional tone}: Clever, uneasy, rebellious, funny, and empowering
\end{promptlist}
\promptsec{Story Core}
\begin{promptlist}
\item \textbf{Characters}: Six recurring characters. Zoe Kim, 26, a contract content moderator with an exceptional eye for patterns; Max Barlow, 35, the company's media-friendly founder; Priya Nair, 30, a data scientist with divided loyalties; Eli Brooks, 24, a livestream creator whose reputation is destroyed by the platform; Ruth Kim, 55, Zoe's practical mother; and NOVA, the platform's recommendation system, represented only through fictional interfaces and synthetic audio.
\item \textbf{Setting}: A trend-forecasting startup, Zoe's small apartment, a creator studio, a late-night diner, and a public product-launch stage.
\item \textbf{One-sentence synopsis}: When a recommendation engine begins predicting private events from her life, a disposable contract worker uncovers a hidden human manipulation team and uses the company's own launch livestream to return control to the people it exploited.
\item \textbf{Pacing requirements}: Construct a clean five-step escalation. Episode 1: an impossible prediction names Zoe before she enters the system. Episode 2: Zoe and Eli trace a pattern linking manufactured scandals to product growth. Episode 3: Priya appears to betray them, and Zoe is publicly blamed for a data leak. Episode 4: the apparent betrayal reveals a lawful evidence-preservation plan, but Max launches NOVA early. Episode 5: Zoe turns the live demonstration into a transparent audit, establishes the human responsibility behind the ``all-knowing'' algorithm, clears Eli, and chooses to build an independent accountability project rather than accept a hollow promotion. Every episode must contain one visually simple discovery and end on a new question, threat, or reversal.
\end{promptlist}
\promptsec{Optional Supplementary Specifications (may be selected partially or in full, with corresponding adjustments as needed)}
\begin{promptlist}
\item \textbf{Character Visual Design}: Zoe wears practical monochrome layers and a green enamel hair clip; Max uses carefully casual luxury clothing; Priya wears geometric earrings that subtly echo network diagrams; Eli's colorful on-camera wardrobe contrasts with his muted off-camera clothes. Preserve interface-free facial consistency and avoid making NOVA a humanoid robot.
\item \textbf{Dialogue Norms}: Use concise contemporary American English with dry humor. Keep dialogue below 250 words per episode. Explain technical ideas through consequences, not jargon. Allow very brief screen text and one recurring line---``Prediction is not permission''---which changes meaning across the series.
\item \textbf{3-Second Opening Hook}: Each episode begins with a personalized NOVA notification that is immediately proven disturbing, misleading, or strategically planted.
\item \textbf{Timing Allocation}: Use 0--5 seconds for the notification hook, 6--30 seconds for context, 31--75 seconds for investigation or conflict, 76--105 seconds for a discovery, and 106--120 seconds for a reversal and cliff-hanger.
\item \textbf{Distribution Specifications}: Supply five distinct episode-cover frames that share one strong visual system. Cover text must be no longer than six words. Keep fictional interface text large, limit each screen to one key data point, and preserve subtitle-safe space.
\item \textbf{Risk Boundaries}: Use no real platform names, logos, or identifiable executives. Do not provide methods for privacy invasion, hacking, or evading security. Keep all evidence collection lawful and non-technical.
\item \textbf{Final Output Checklist}: Deliver five finished episodes, a series cover, episode titles, a 75-word description, six tags, full subtitle text, fictional interface copy, and one next-episode teaser after Episodes 1--4.
\end{promptlist}
\promptsec{Delivery Requirements}
\promptpar{I will only provide the objective and acceptance criteria; the execution details are fully delegated to you.}
\begin{promptlist}
\item \textbf{Objective}: Deliver a finalized five-episode vertical short-drama series ready for direct distribution, aligned with the above specifications.
\item \textbf{Acceptance criteria}: The series must make its ethical conflict understandable to non-technical viewers, provide fair clues before each reversal, complete Zoe's counterattack arc, maintain interface and character continuity, and deliver finished videos rather than drafts or concept materials.
\end{promptlist}
\end{promptbox}

\Needspace{12\baselineskip}
\begin{promptbox}{Brief 4: The Nineteenth Floor}{5 $\times$ 240\,s \quad 16:9}
\promptpar{You are a professional short drama director and screenwriter. I intend to produce a short drama video with the requirements outlined below. Please engage in consultation with me regarding any ambiguous or uncertain aspects of the project.}
\promptsec{Basic Specifications}
\begin{promptlist}
\item \textbf{Format}: Multi-episode series (5 episodes)
\item \textbf{Single-episode duration}: 4 minutes
\item \textbf{Aspect ratio}: 16:9 (horizontal screen)
\item \textbf{Production model}: Text-based story development, image-generated visual assets, and video-based final production
\end{promptlist}
\promptsec{Content Positioning}
\begin{promptlist}
\item \textbf{Genre}: Apartment-building mystery, social suspense, and ensemble emotional drama
\item \textbf{Target audience}: Adult streaming viewers, mystery audiences, urban residents, and fans of character-driven limited series
\item \textbf{Overall style}: Cinematic, claustrophobic, realistic, clue-based, and strongly atmospheric
\item \textbf{Emotional tone}: Suspenseful, morally complex, humane, and cautiously hopeful
\end{promptlist}
\promptsec{Story Core}
\begin{promptlist}
\item \textbf{Characters}: Seven recurring characters. Mara Ellis, 34, a paramedic on leave who has stopped trusting her judgment; Theo Grant, 17, a highly observant student and amateur sound editor; Mrs. Alvarez, 65, the building's unofficial historian; Simon Pike, 45, the polished property manager; Nia Okafor, 39, a tenant-rights lawyer; Caleb Ross, 30, a night security guard; and Anna Vale, 28, a tenant who has apparently vanished from Apartment 1904.
\item \textbf{Setting}: A high-rise apartment building during one storm-related overnight lockdown, including the lobby, elevators, stairwell, security room, mechanical corridor, roof, and the sealed nineteenth floor.
\item \textbf{One-sentence synopsis}: During a blackout, residents hear a missing tenant calling through an emergency speaker from a floor that officially does not exist, forcing a disgraced paramedic and a teenage sound editor to uncover an illegal hidden residence before the building manager erases the evidence at dawn.
\item \textbf{Pacing requirements}: Treat each episode as one stage of a fair-play mystery. Episode 1 establishes the voice, the impossible floor indicator, and Anna's disappearance; it ends when Mara receives a live message from Apartment 1904. Episode 2 maps contradictory building plans, seeds each resident's motive, and ends with security footage showing Mara entering the sealed floor earlier that night. Episode 3 reveals that the footage timestamp was altered and that hidden rooms were used to house vulnerable tenants off the lease; it ends when Theo disappears. Episode 4 finds Theo in the mechanical corridor, reveals Caleb as a frightened witness rather than the culprit, and lets Simon trigger an evacuation to destroy the records. Episode 5 turns the evacuation into a coordinated rescue and evidence handoff, finds Anna alive in a service refuge, explains every planted clue, and gives Mara a credible step toward recovering her confidence. Each episode needs a midpoint reversal and a final cliff-hanger, while Episode 5 must provide complete resolution.
\end{promptlist}
\promptsec{Optional Supplementary Specifications (may be selected partially or in full, with corresponding adjustments as needed)}
\begin{promptlist}
\item \textbf{Character Visual Design}: Give each character a readable silhouette and one practical light source during the blackout. Mara carries a compact medical flashlight, Theo wears silver over-ear headphones around his neck, Mrs. Alvarez uses an old green camping lantern, and Simon carries a pristine white emergency torch. Wardrobe must remain consistent across the single night.
\item \textbf{Shot and Visual Composition}: Use negative space, long corridors, foreground obstructions, security-camera angles, and slow creeping movement. Darkness must retain visible facial detail and navigable spatial information. Reserve handheld camera for genuine danger, and use stable compositions when a character tells the truth.
\item \textbf{Dialogue Norms}: Use restrained contemporary English, overlapping ensemble dialogue only when intelligible, and no more than 500 spoken words per episode. Avoid characters explaining clues they already know. Use building announcements sparingly and style them distinctly in subtitles.
\item \textbf{Sound Design}: Sound is part of the evidence. Establish consistent acoustic signatures for the elevator motor, rooftop wind, water pipes, emergency speaker, and service lift. Theo's deductions must arise from audible details already presented to the audience. Do not use false stingers that contradict the story reality.
\item \textbf{3-Second Opening Hook}: Begin each episode with a different fragment of Anna's voice over a black screen, followed by an image that complicates the meaning of her words.
\item \textbf{Timing Allocation}: For each four-minute episode, use 0--8 seconds for the hook, 9--55 seconds for immediate stakes, 56--130 seconds for investigation and interpersonal conflict, 131--190 seconds for the midpoint reversal, 191--225 seconds for escalating consequence, and 226--240 seconds for the cliff-hanger or final resolution.
\item \textbf{Continuity Rules}: Maintain an explicit geography of the building, consistent emergency-light colors, plausible travel times, storm progression, phone battery levels, injuries, wet clothing, key-card possession, and the exact chronology shown by clocks and recordings.
\item \textbf{Distribution Specifications}: Supply horizontal and vertical-safe cover crops, five episode title cards, and legible clue inserts. Avoid crushed blacks, flashing light patterns, and dense architectural diagrams that cannot be understood quickly.
\item \textbf{Risk Boundaries}: Avoid graphic harm, procedural instructions for bypassing building security, stigmatizing vulnerable residents, and references to real property companies. Emergency care must be depicted responsibly and without unrealistic guarantees.
\item \textbf{Final Output Checklist}: Deliver five finished episodes, series and episode covers, titles, a spoiler-free synopsis, tags, complete subtitles, an internal continuity map, and teaser lines for Episodes 1--4.
\end{promptlist}
\promptsec{Delivery Requirements}
\promptpar{I will only provide the objective and acceptance criteria; the execution details are fully delegated to you.}
\begin{promptlist}
\item \textbf{Objective}: Deliver a finalized five-episode horizontal mystery series ready for direct distribution, aligned with the above specifications.
\item \textbf{Acceptance criteria}: The mystery must be solvable from visible and audible clues, the building geography and single-night timeline must be coherent, every principal character must serve the plot, all major questions must be answered, and the output must consist of complete finished episodes rather than drafts.
\end{promptlist}
\end{promptbox}

\Needspace{12\baselineskip}
\begin{promptbox}{Brief 5: The Last Clause}{6 $\times$ 300\,s \quad 9:16}
\promptpar{You are a professional short drama director and screenwriter. I intend to produce a short drama video with the requirements outlined below. Please engage in consultation with me regarding any ambiguous or uncertain aspects of the project.}
\promptsec{Basic Specifications}
\begin{promptlist}
\item \textbf{Format}: Multi-episode series (6 episodes)
\item \textbf{Single-episode duration}: 5 minutes
\item \textbf{Aspect ratio}: 9:16 (vertical screen)
\item \textbf{Production model}: Text-based story development, image-generated visual assets, and video-based final production
\end{promptlist}
\promptsec{Content Positioning}
\begin{promptlist}
\item \textbf{Genre}: Inheritance mystery, family power struggle, legal drama, and female-led revenge thriller
\item \textbf{Target audience}: Adult drama viewers, fans of family intrigue and high-stakes comeback stories, and mobile audiences who enjoy layered mysteries
\item \textbf{Overall style}: Premium, tense, elegant, twist-heavy, emotionally grounded, and cathartic
\item \textbf{Emotional tone}: Suspicious, controlled, painful, defiant, and ultimately liberating
\end{promptlist}
\promptsec{Story Core}
\begin{promptlist}
\item \textbf{Characters}: Eight recurring characters. Ava Mercer, 36, a meticulous probate lawyer who left her wealthy family fifteen years ago; Helena Mercer, 62, Ava's image-conscious stepmother and acting head of the family foundation; Marcus Mercer, 39, Ava's impulsive half-brother; Lily Mercer, 22, Marcus's idealistic daughter and a journalism student; Samuel Price, 50, the family's longtime solicitor; Dr. Evan Cole, 37, Ava's former friend and the late patriarch's home-care coordinator; Rosa Bell, 64, the estate housekeeper and keeper of its oral history; and Thomas Mercer, Ava's deceased father, present through legally recorded messages and personal objects.
\item \textbf{Setting}: A secluded modern estate, its archive room, greenhouse, family foundation office, probate hearing room, lakeside boathouse, and formal dining hall over six consecutive days.
\item \textbf{One-sentence synopsis}: Summoned home to read the will of the father who disowned her, a probate lawyer learns that she will inherit nothing unless she proves which member of her family fabricated the final clause---and discovers that the clause was designed to expose a deeper theft from the charitable foundation.
\item \textbf{Pacing requirements}: Build one evidence-led movement per episode while preserving a continuous emotional arc. Episode 1 must establish the hostile reunion, play Thomas's first recorded message, and reveal that Ava herself is named as the suspected forger. Episode 2 must test the document trail, expose Marcus's debts, and end with a second will hidden inside the greenhouse irrigation ledger. Episode 3 must show that the second will is a decoy, reveal diverted foundation funds, and end when Samuel produces footage suggesting Ava met Thomas on the night the clause changed. Episode 4 must reconstruct that meeting, uncover Helena's pressure campaign, and end with Lily publishing incomplete evidence that turns the public against Ava. Episode 5 must force Ava to choose between clearing her name and protecting Lily, reveal Samuel as the architect of the financial scheme, and end when the original signed codicil appears to destroy Ava's case. Episode 6 must establish how the authentic signature was obtained without providing replicable forgery techniques, use Rosa's and Evan's testimony to complete the timeline, expose Samuel while giving Helena and Marcus morally distinct consequences, restore the foundation's stolen funds through credible legal action, and let Ava reject personal control of the estate in favor of independent oversight. Every episode must include one plausible clue, one emotionally costly choice, one reversal near the final minute, and a strong closing image.
\end{promptlist}
\promptsec{Optional Supplementary Specifications (may be selected partially or in full, with corresponding adjustments as needed)}
\begin{promptlist}
\item \textbf{Character Visual Design}: Ava wears structured dark clothing that gradually incorporates her late mother's deep blue; Helena wears immaculate ivory and gold; Marcus appears expensive but slightly disordered; Lily carries a small red field recorder; Samuel uses a dark green fountain pen; Rosa wears a ring of old brass keys. Preserve age, facial features, wardrobe progression, and object ownership across all six episodes.
\item \textbf{Shot and Visual Composition}: Use formal family tableaux, shallow-focus document inserts, reflections in glass architecture, controlled dolly movement, and occasional overhead compositions that turn the dining table into a strategic board. The greenhouse should begin lush and deceptive, then become visually transparent in the final episode. Keep all document text fictional, selective, and legible.
\item \textbf{Dialogue Norms}: Use precise contemporary English with distinct registers for legal, family, and private speech. Keep each episode below 650 spoken words. Legal concepts must be understandable through conflict and consequence, not lectures. Recorded messages from Thomas may not exceed 70 words per episode. Avoid constant one-line quips; allow silence after major betrayals.
\item \textbf{Sound Design}: Use restrained low strings, lake ambience, greenhouse sprinklers, the scratch of a fountain pen, archive-box movement, and a recurring three-note piano phrase tied to Ava's mother. Reduce the score during testimony and evidence reveals so spoken details remain clear.
\item \textbf{3-Second Opening Hook}: Each episode begins with a legally or emotionally explosive fragment---a name in the will, a hidden signature, a transfer total, a security timestamp, a published accusation, or a final recorded instruction---before returning to the immediate cause.
\item \textbf{Timing Allocation}: For each five-minute episode, use 0--10 seconds for the hook, 11--65 seconds for orientation and immediate stakes, 66--165 seconds for investigation and family conflict, 166--230 seconds for the central confrontation, 231--275 seconds for a revelation and its consequence, and 276--300 seconds for the cliff-hanger or final resolution.
\item \textbf{Continuity Rules}: Maintain a master chronology for all six days, document versions, signatures, recordings, key possession, financial clues, weather, injuries, costumes, meals, and character knowledge. No character may act on information they have not yet received.
\item \textbf{Distribution Specifications}: Provide one series cover and six episode covers with a consistent estate-and-document motif. Cover copy must remain under seven words. Maintain vertical composition, facial readability, subtitle safety, bright-enough night scenes, and one dominant clue per frame.
\item \textbf{Risk Boundaries}: All legal procedures, institutions, charities, companies, and documents must be fictional. Avoid real-person resemblance, actionable forgery or financial-crime methods, defamatory parallels, graphic violence, and simplistic portrayals of illness or capacity. The resolution must rely on lawful evidence and credible due process rather than vigilante punishment.
\item \textbf{Final Output Checklist}: Deliver six finished episodes, a series cover, six episode covers, three series-title options, episode titles, a 100-word spoiler-free synopsis, eight tags, full subtitle text, all fictional document and screen copy, continuity notes, and next-episode teasers for Episodes 1--5.
\end{promptlist}
\promptsec{Delivery Requirements}
\promptpar{I will only provide the objective and acceptance criteria; the execution details are fully delegated to you.}
\begin{promptlist}
\item \textbf{Objective}: Deliver a finalized six-episode vertical family mystery ready for direct distribution, aligned with the above specifications.
\item \textbf{Acceptance criteria}: The inheritance and financial mysteries must be internally coherent and fairly clued, legal developments must feel credible without becoming procedural exposition, every major character must receive a meaningful consequence, the emotional and plot arcs must conclude in Episode 6, and the delivery must consist of finished videos rather than drafts, outlines, or sample footage.
\end{promptlist}
\end{promptbox}

\end{document}